\documentclass[runningheads]{llncs}
\usepackage{graphicx}
\usepackage{tikz}
\usepackage{comment}
\usepackage{amsmath,amssymb} % define this before the line numbering.
\usepackage{color}

\usepackage[accsupp]{axessibility}  % Improves PDF readability for those with disabilities.

\usepackage{AbdAlmageed_style}
\usepackage{Jiazhi_style}
\usepackage{subcaption}
\usepackage{wrapfig}
\usepackage{subfiles}
\usepackage{color,soul}
\usepackage{xcolor}
\usepackage{multirow}
\usepackage{bm}
\usepackage{makecell}
\usepackage[accsupp]{axessibility}
\usepackage{algorithm}
\usepackage{algpseudocode}
\usepackage{tabularx,booktabs}
\newcolumntype{Y}{>{\centering\arraybackslash}X}
\usepackage{orcidlink}

\begin{document}
% \renewcommand\thelinenumber{\color[rgb]{0.2,0.5,0.8}\normalfont\sffamily\scriptsize\arabic{linenumber}\color[rgb]{0,0,0}}
% \renewcommand\makeLineNumber {\hss\thelinenumber\ \hspace{6mm} \rlap{\hskip\textwidth\ \hspace{6.5mm}\thelinenumber}}
% \linenumbers
\pagestyle{headings}
\mainmatter
\def\ECCVSubNumber{51}  % Insert your submission number here

\title{CAT: Controllable Attribute Translation for Fair Facial Attribute Classification}
% RAFT: Recurrent All-Pairs Field Transforms for
% Optical Flow
% Synthetic dataset for multi attributes classification} % Replace with your title

% INITIAL SUBMISSION 
% %\begin{comment}
% \titlerunning{ECCV-22 submission ID \ECCVSubNumber} 
% \authorrunning{ECCV-22 submission ID \ECCVSubNumber} 
% \author{Anonymous ECCV submission}
% \institute{Paper ID \ECCVSubNumber}
% %\end{comment}
%******************

% CAMERA READY SUBMISSION
% \begin{comment}
\titlerunning{CAT}
% If the paper title is too long for the running head, you can set
% an abbreviated paper title here
%

\author{Jiazhi Li\inst{1,2,3}\orcidlink{0000-0003-3938-7989} \and Wael Abd-Almageed\inst{1,2,3}\orcidlink{0000-0002-8320-8530}}
\authorrunning{J. Li and W. Abd-Almageed}
% First names are abbreviated in the running head.
% If there are more than two authors, 'et al.' is used.
%
\institute{USC Ming Hsieh Department of Electrical and Computer Engineering \and
USC Information Sciences Institute \and Visual Intelligence and Multimedia Analytics Laboratory\\
\email{\{jiazli, wamageed\}@isi.edu}}
%******************
\maketitle

% \author{First Author\inst{1}\orcidID{0000-1111-2222-3333} \and
% Second Author\inst{2,3}\orcidID{1111-2222-3333-4444} \and
% Third Author\inst{3}\orcidID{2222--3333-4444-5555}}
% Jiazhi 
% % 0000-0003-3938-7989

% wael 
% % 0000-0002-8320-8530

%%%%%%%%% ABSTRACT
\begin{abstract}
As the social impact of visual recognition has been under scrutiny, several \emph{protected-attribute} balanced datasets emerged to address \emph{dataset bias} in imbalanced datasets. However, in facial attribute classification, \emph{dataset bias} stems from both \emph{protected attribute} level and facial attribute level, which makes it challenging to construct a multi-attribute-level balanced real dataset. To bridge the gap, we propose an effective pipeline to generate high-quality and sufficient facial images with desired facial attributes and supplement the original dataset to be a balanced dataset at both levels, which theoretically satisfies several fairness criteria. The effectiveness of our method is verified on sex classification and facial attribute classification by yielding comparable task performance as the original dataset and further improving fairness in a comprehensive fairness evaluation with a wide range of metrics. Furthermore, our method outperforms both resampling and balanced dataset construction to address \emph{dataset bias}, and debiasing models to address \emph{task bias}.
 
\end{abstract}

%%%%%%%%% BODY TEXT
\section{Introduction}
\label{introduction}
Bias issues in machine learning methods involving \emph{protected attributes}~\cite{sex_race_PPB, race2} (\eg sex and race) have lately garnered tremendous attention~\cite{FRVT3,FR_bias_study,AI_fairness_360}, such as in facial attribute classification~\cite{debias_domain_independent_training_BA_usage, hat_glasses_correlation} and sex classification~\cite{sex_race_PPB} on face dataset. Many studies~\cite{debias_domain_independent_training_BA_usage, sex_race_PPB} show that one of the most obvious phenomena of bias issues in these downstream tasks is that the underlying learning algorithms yield uneven performance for different demographic groups and relatively worse performance for minorities. Such unfair performance across different demographic cohorts has hampered credibility of computer vision algorithms on face dataset from both individuals and the whole society, which is urgently to be resolved.

Ensuring fairness of visual recognition on face dataset, has aroused great interest in the machine learning~\cite{debias_resampling1, debias_cost_sensitive, debias1_no_bam_DP, debias2_ACC, debias3_ba2, debias_adversarial_forgetting_yz,debias_transfer_learning,debias_multi_task,debias_age}. Without loss of generality, there are two main challenges with respect to two distinct types of bias --- (1) lack of generalized learned representations due to the spurious correlation between prediction target and confounding factors, which could include sensitive attribute (\eg sex) in training dataset, referred to as \emph{sensitive attribute bias} and \emph{task bias}~\cite{debias_adversarial_uniform_confusion_ACC_unbalanced_dataset2_LAOFIW,cross_sample_MI_minimization, End}, and (2) uneven performance across cohorts caused by insufficient samples for particular demographic groups in the existing dataset, referred to as \emph{minority group bias} or \emph{dataset bias}~\cite{def_dataset_bias, balanced_dataset_BFW, def_minority_bias}.
 
\begin{wrapfigure}{r}{0.5\textwidth}
\begin{center}
   \includegraphics[width=0.8\linewidth]{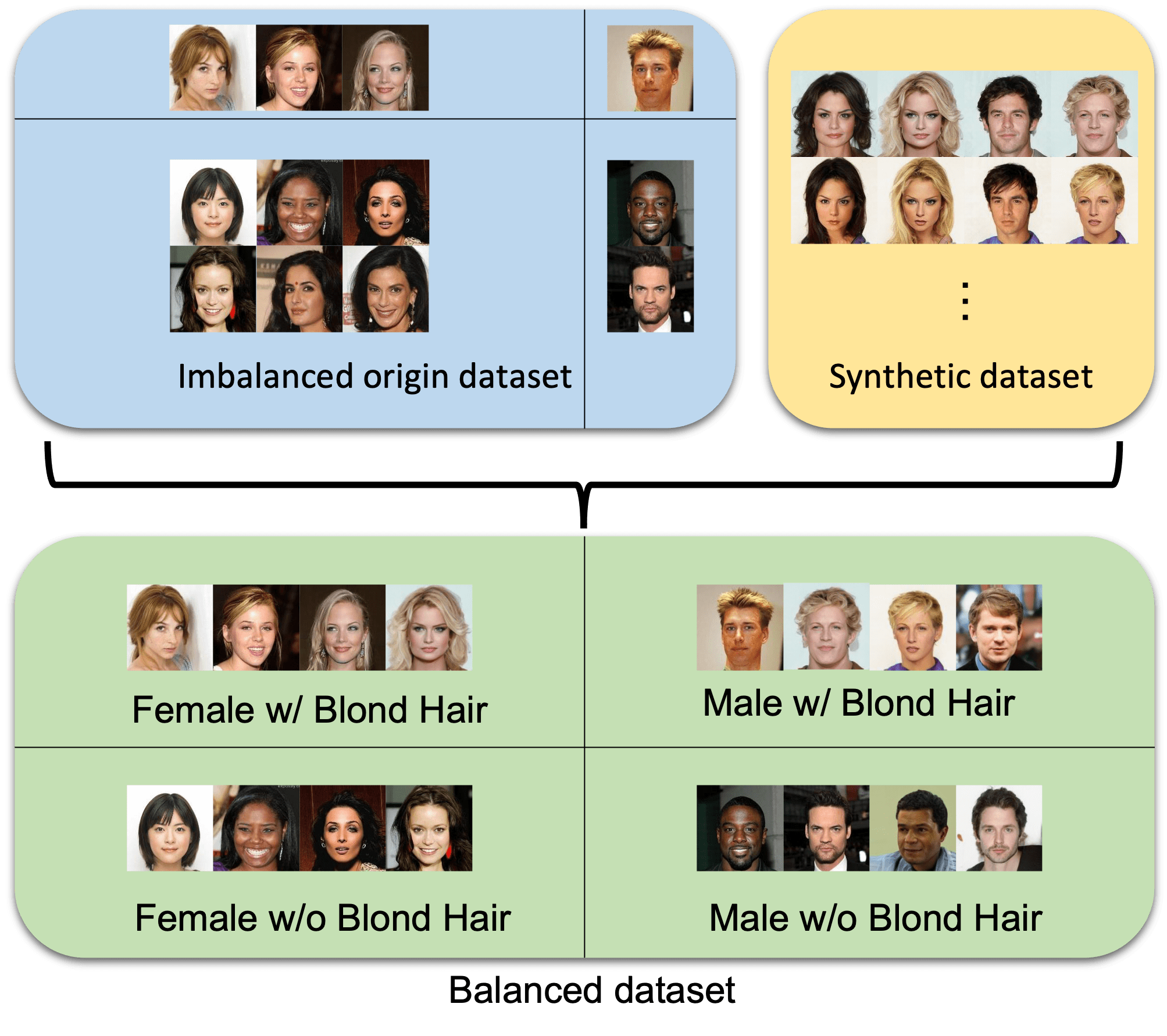}
\end{center}
   \caption{As the original facial attribute dataset may be imbalanced at both \emph{protected attribute} level (\eg sex) and facial attribute level (\eg blond hair), which induce \emph{dataset bias}, we propose a pipeline to generate high-quality synthetic datasets with sufficient samples in a balanced distribution of both levels to rectify skew of the original dataset.}
\label{fig:idea}
\end{wrapfigure}
 
Mainstream debiasing methods~\cite{cross_sample_MI_minimization,End,debias4_extreme_bias_MI_colored_MNIST} focus on mitigating~\emph{task bias}, with less emphasis on tackling \emph{dataset bias}. Furthermore, although these methods effectively address \emph{task bias}~\cite{debias_domain_independent_training_BA_usage}, they are limited in addressing the issue of insufficient samples of minority demographics in the training dataset, as demonstrated in~\cref{tab:comparison} with high bias scores. Instead, in the specific scenarios involving \emph{dataset bias}, constructing a balanced dataset is a more appropriate way. For example, some work mitigate sex distribution differences or race distribution differences by either collecting fair dataset~\cite{FairFace,UTKFace} or strategically resampling from existing datasets~\cite{debias_resampling1, debias_cost_sensitive}. However, merely maintaining the balance across the \emph{protected attributes} may not resolve the risk of \emph{dataset bias} which may be implicitly derived from imbalance on other facial attributes (\eg blond hair or chubby)~\cite{hat_glasses_correlation}. Furthermore, due to the recourse limitations in academia to collect sufficient real images for minority groups, invasion of privacy while collecting more real data and the impossibility of covering all known and unknown confounding factors, it may be impractical to construct a balanced real dataset which is balanced across both \emph{protected attributes} and all facial attributes so as to raise the research scenario to address \emph{dataset bias}.
 
We therefore propose a method to iteratively generate high-quality synthetic images with desired facial attributes and combine the synthetic dataset with the original dataset to create a semi-synthetic balanced dataset by supplementing insufficient samples in both \emph{protected attribute} level and facial attribute level for downstream tasks, as illustrated in~\cref{fig:idea}. We use facial attribute classification and sex classification on face dataset as the proxy problem to discuss the effectiveness of our method instead of face recognition since the generation of synthetic face recognition ground truth relies on pre-trained models~\cite{synthetic_data_face_recognition, synthetic_data_face_recognition2}, which may be already biased involving \emph{protected attributes} in debiasing face recognition literature. On the other hand, the generation of facial attribute ground truth of our synthetic dataset is bundled with synthetic images directly rather than rely on classifiers, as elaborated in~\cref{sec:method}. To construct semi-synthetic datasets which are balanced at both \emph{protected attribute} level and facial attribute level, our method explicitly controls the facial attributes during image generation. Compared with debiasing models that focus on the process to learn fairly, our method to investigate training dataset to achieve fairness is super straightforward and effective, which is easier for generalization. Furthermore, compared with fair dataset collection restricted by long-tail distribution~\cite{Long_tail} and limited resources, the advantages of semi-synthetic dataset allow our method to generate sufficient face images with more diversity of facial attributes. Finally, different from data augmentation methods by adding slightly modified copies in existing datasets, we create a generative pipeline to first learn the facial attribute appearance of real data and then generate synthetic images with desired facial attributes. The key contributions of this paper can be summarized as follows:
 
\begin{itemize}
    \item A pipeline to construct semi-synthetic face datasets to introduce fairness in facial attribute classification and sex classification.
    \item An investigation of the nature of \emph{dataset bias} stemming from the facial attribute-level imbalance instead of the \emph{protected attribute}-level imbalance.
    \item A comprehensive fairness evaluation for a wide range of debiasing techniques.
\end{itemize}
\section{Related Work}
\label{sec:related_work}

\noindent
\textbf{Debiasing face datasets.}
Long tail distribution~\cite{Long_tail} leads to inequity towards minorities in face datasets~\cite{CelebA, IJBC}, and several methods have been designed to address \emph{dataset bias} by collecting fair datasets, such as PPB~\cite{PPB_metric_Joy}, RFW~\cite{debias_domain_discriminative_RFW_MI_AUC_TAR_FAR}, UTKFace~\cite{UTKFace} and FairFace~\cite{FairFace}. Meanwhile, strategic resampling methods~\cite{RL-RBN,debias_resampling1, debias_cost_sensitive} attempt to balance the \emph{appearance} of training data with respect to different demographic groups, referred to as \emph{domain}. Since fair dataset and strategic resampling address bias issues before model training, in~\cite{preprocessing1,preprocessing2}, they are also referred as pre-processing methods. On the other hand, to mitigate \emph{task bias} from the perspective of model training, some debiasing models have been proposed. First, adversarial forgetting methods~\cite{debface_representation_disentanglement,debias_adversarial_uniform_confusion_ACC_unbalanced_dataset2_LAOFIW} manage to extract a new representation which only contains information for the recognition task and exclude the information of \emph{protected attributes}. Second, domain adaptation methods~\cite{debias_domain_discriminative_RFW_MI_AUC_TAR_FAR,Image_caption_Bias_amplification_yz} adapt learned recognition knowledge from the majority \emph{domain} to the minority \emph{domain}. Third, domain independent training methods~\cite{GAC, debias_domain_independent_training_BA_usage} delicately design several separate classifiers in different demographic groups and garner an ensemble of classifiers by \emph{representation sharing}.

\noindent
\textbf{Fairness using GANs.}
Over the past couple years, Generative Adversarial Network (GAN)~\cite{GAN} is developed further in a tremendous number of work~\cite{gan_development1,gan_development2,gan_development3,gan_development4,gan_development5,gan_development6}. As pointed by~\cite{GANs_Objectives}, three main purposes of GAN are fidelity, diversity and privacy. Beside the mainstream developments, GAN has also been used to augment real datasets~\cite{gan_augment_real_data1_low_shot,gan_augment_real_data2_long_tail,gan_augment_real_data3}. Some work~\cite{infogan, bidirectional_gan} manipulate latent vectors to augment real datasets to learn disentangled representations. Moreover, Balakrishnan \etal~\cite{transect} proposes an experimental method with face images generated by StyleGAN2~\cite{styleGan2} to reveal correlation between \emph{protected attributes} and fairness performance, which in nature is to study and measure bias instead of debiasing. While~\cite{DECAF, FairGAN,fairgan+} use GANs to construct synthetic dataset to improve fairness, they conduct experiments on Census dataset, \eg Adult dataset (predict whether income exceeds \$50K/yr based on census data), which is inapplicable for face dataset since label and data of census dataset are both census data and can be generated together, which is different from generating accurate labels together with generated images in face dataset.

Recently, a minority route of augmentation by GANs to introduce fairness~\cite{hat_glasses_correlation,generate_fair2_fairness_GAN, generate_fair3} emerged. However, as pointed out by~\cite{hat_glasses_correlation}, some work~\cite{generate_fair1,generate_fair2_fairness_GAN, generate_fair3} need to train a new GAN per bias since different facial attributes may yield different levels of bias. By contrast, V. Ramaswamy \etal~\cite{hat_glasses_correlation} use a single GAN to construct a fair synthetic dataset by generating the complementary image with the same target label but the opposite \emph{protected attribute} label corresponding to the randomly generated image. However, to assign the attribute labels, \cite{hat_glasses_correlation} relies on an attribute classifier trained on the original dataset, which may be harmful for fairness since the pre-trained classifier may be already biased by skew of the original dataset. By contrast, our method does not need the pre-trained classifier or significant resources of human annotations using Amazon Mechanical Turk as in~\cite{transect} to assign labels for each generated image. As elaborated in~\cref{sec:method}, our method only needs a small set of images with labels as seeds and then construct the whole synthetic datasets automatically with labels, which can be directly used in the following performance and fairness evaluation pipeline. Furthermore, compared with~\cite{hat_glasses_correlation} using additional images for the good recognition performance, our method achieves comparable performance and better fairness even with the same-size training dataset, as elaborated in~\cref{subsec:ablation_study}. 
\section{Approach}
\label{sec:method}

As discussed in~\cref{sec:related_work}, some methods~\cite{generate_fair1,generate_fair2_fairness_GAN, generate_fair3} focus on applying different data augmentation techniques to mitigate different types of bias. By contrast, we train a unified GAN on the original dataset to introduce fairness for both \emph{protected attributes} and all facial attributes. 

\begin{algorithm}[t]
\caption{\emph{Controllable Attribute Translation} (CAT).}
\label{alg:similarity}
\begin{algorithmic}[1]
 \State \textbf{initialize} $A^{intra}_Y$ and $B^{inter}_Y$ to be an empty set.
 \For{\texttt{$e_i \in \mathbb{R}^k$, $e_i \in S_{Y}$}}
    \For{\texttt{$e_j \in \mathbb{R}^k$, $e_j \in S_{Y}$}} 
        \State $A_{ij} = \{l~|~|e^l_i - e^l_j| < \texttt{intra\_threshold}, l\in[1,k]\}$
        \State $A^{intra}_{Y} \leftarrow A^{intra}_{Y} \cap A_{ij}$ \Comment{Intra-class similarity.}
    \EndFor
    \For{\texttt{$\bar{e}_j \in \mathbb{R}^k$, $\bar{e}_j \in S_{\bar{Y}}$}}
        \State $B_{ij} = \{l~|~|e^l_i - \bar{e}^l_j| > \texttt{inter\_threshold}, l\in[1,k]\}$
        \State $B^{inter}_{Y} \leftarrow B^{inter}_{Y} \cap B_{ij}$ \Comment{Inter-class difference.}
    \EndFor
 \EndFor
 \State \textbf{output} $C_Y \leftarrow A_Y^{intra} \cup B_Y^{inter}$
\end{algorithmic}
\end{algorithm}

In the facial attribute classification task, given an attribute dataset $\mathcal{D}$ involving instances $(x_i,y_i,z_i)$, the classification model $H$ consumes an image $x_i \in \mathcal{X}$ annotated with a set of binary facial attributes $y_i \in \mathcal{Y}$ (\eg hair color) and \emph{protected attributes} $z_i \in \mathcal{Z}$ (\eg sex), and produces the predicted facial attribute label $y_i' \in \mathcal{Y}$. As illustrated by conditions of three fairness criteria~\cite{DP} defined with conditional probability, \emph{dataset bias} may be induced from skew of training dataset. Specifically, $\mathcal{D}$ may naturally be imbalanced across the positive samples $Z$ and the negative samples $\bar{Z}$ with respect to the \emph{protected attribute} $\mathcal{Z}$ (if binary), \ie $|\mathcal{D}_{Z}| \neq |\mathcal{D}_{\bar{Z}}|$, on which the classification model $H$ is trained may invalidate \emph{demographic parity}~\cite{DP} which requires that the prediction label $Y'$ (if binary) and the \emph{protected attribute} $\mathcal{Z}$ are independent, \ie
\begin{align}
    \label{DP}
    P(Y' = 1) = P(Y' = 1|Z) = P(Y' = 1|\bar{Z}).
\end{align}

Furthermore, the original dataset $\mathcal{D}$ may be imbalanced at the facial attribute level across \emph{protected attributes} for the facial attribute $\mathcal{Y}$, \ie $|\mathcal{D}_{Z}^{Y}| \neq |\mathcal{D}_{\bar{Z}}^{Y}|$, where $|\mathcal{D}_{Z}^{Y}|$ is the number of positive samples $Y$ among $Z$ and $|\mathcal{D}_{\bar{Z}}^{Y}|$ is the number of $Y$ among $\bar{Z}$ with respect to the facial attribute $\mathcal{Y}$, which invalidates \emph{equal opportunity}~\cite{DP}, an another criterion that requires independence between the prediction label $Y'$ and the \emph{protected attribute} $\mathcal{Z}$ conditional on $Y$, \ie
\begin{align}
    \label{EO}
    P(Y' = 1| Z, Y) = P(Y' = 1| \bar{Z}, Y).
\end{align}
Finally, \emph{equalized odds}~\cite{DP}, which is a stronger criterion, requires that $Y'$ and $\mathcal{Z}$ are independent conditional on both $Y$ and $\bar{Y}$, \ie with the additional constraint than \emph{equal opportunity} as followed,
\begin{align}
    \label{EOS}
     P(Y' = 1| Z, \bar{Y}) = P(Y' = 1| \bar{Z}, \bar{Y}),
\end{align}
where $\bar{Y}$ represents the negative samples with respect to the facial attribute $\mathcal{Y}$. To mitigate the influence from skew of training dataset under the requirements of \emph{equalized odds}, $\mathcal{D}$ should be balanced across $\mathcal{Z}$ for both $Y$ and $\bar{Y}$, \ie $|\mathcal{D}_{Z}^{Y}| = |\mathcal{D}_{\bar{Z}}^{Y}|$ and $|\mathcal{D}^{\bar{Y}}_{Z}| = |\mathcal{D}^{\bar{Y}}_{\bar{Z}}|$, where $|\mathcal{D}_{Z}^{\bar{Y}}|$ is the number of $\bar{Y}$ among $Z$ and $|\mathcal{D}_{\bar{Z}}^{\bar{Y}}|$ is the number of $\bar{Y}$ among $\bar{Z}$ with respect to the facial attribute $\mathcal{Y}$.

However, even though prior methods that depend on fair dataset collection~\cite{FairFace} or resampling~\cite{debias_resampling1, debias_cost_sensitive} may satisfy \emph{demographic parity} by balancing dataset across \emph{protected attributes}, it is hard for these methods to satisfy \emph{equal opportunity}  and/or \emph{equalized odds} constraints which require balance at the facial attribute level due to the lack of sufficient images which are balanced across one facial attribute, much less across multiple facial attributes in the multi-attribute classification task. Thus, to mitigate the effect of \emph{dataset bias} on the trained model, we generate a semi-synthetic dataset $\mathcal{D}_{syn}$ which meets requirements of all three fairness criteria by controllably translating the representation of facial attributes in latent space to image space and generate sufficient images with desired facial attributes, referred to as \emph{Controllable Attribute Translation} (CAT).

Given a $k$-dimensional latent space $\mathcal{E} \in \mathbb{R}^k$, a generative model $G: \mathcal{E} \to \mathcal{I}$ trained on real dataset $\mathcal{D}$ produces the synthetic image $I_i$ from the latent vector $e_i \in \mathcal{E}$. By human annotations denoted as $F$, we construct a \textbf{succinct} set of latent vectors with the facial attribute $\mathcal{Y}$, referred to as \emph{attribute seeds} $S_{Y} \subset \mathcal{E}$, such that $\forall e_i \in S_{Y}, F(G(e_i)) = Y$. We refer to $\mathcal{Y}$ as Attribute of Interest (AOI)\footnote{AOI is set based on interests for different experiments.}.

To stably generate images with AOI, we first study the common similarity of latent vectors among $S_{Y}$ since $I_i$ may yield same AOI and randomly yield other facial attributes, where $I_i = G(e_i), e_i \in S_{Y}$. \jiazhi{Given two randomly picked latent vectors $e_i, e_j \in S_{Y}$, we traverse and aggregate the dimension indices where the absolute difference is smaller than a hyperparameter \texttt{intra\_threshold}, \ie $ A_{ij} = \{l~|~|e^l_i - e^l_j| < \texttt{intra\_threshold}, l\in[1,k]\}$, which are the dimensions representing the attribute similarity between $e_i$ and $e_j$. Furthermore, since pairs of images generated by latent vectors from $S_Y$ may be similar in other unwanted facial attributes, to purify the similarity of AOI and find predominate dimensions controlling AOI, we traverse all combination of pairs in $S_Y$ to find $A_{ij}$ of each pair and take the intersection across all $A_{ij}$, \ie $A^{intra}_{Y} = \bigcap_{e_i,e_j \in S_Y} A_{ij}$, which is the least common similarity, referred to as \emph{intra-class similarity} of AOI, which effectively eliminates the similarity of other unwanted facial attributes.}

\begin{figure}[t]
\begin{center}
  \includegraphics[width=0.96\linewidth]{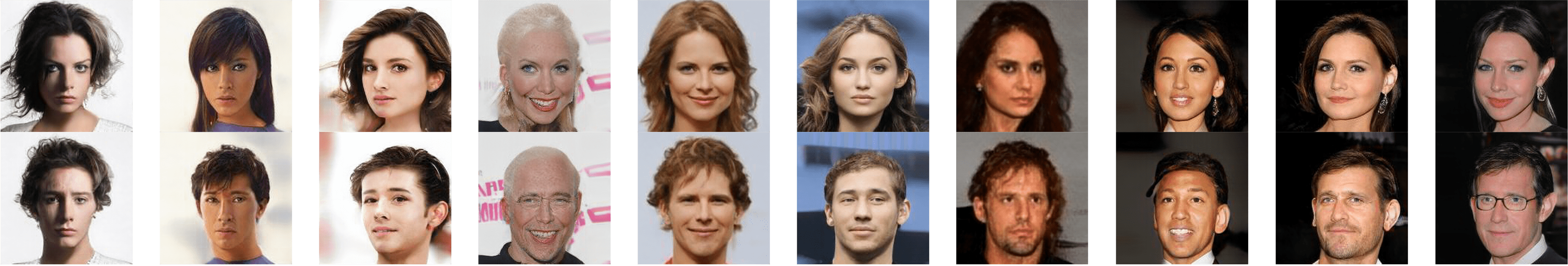}
\end{center}
  \caption{Examples of generated female images and male images.}
\label{fig:sex}
\end{figure}

To select more representative dimension indices for AOI, we compensate \emph{intra-class similarity} with \emph{inter-class difference}. We randomly pick $e_i \in S_{Y}$ and $\bar{e_i} \in S_{\bar{Y}}$, where $S_{\bar{Y}}$ is the set of latent vectors without AOI, such that $\forall \bar{e}_i \in S_{\bar{Y}}, F(G(\bar{e}_j)) = \bar{Y}$. \jiazhi{Further, we aggregate the dimension indices where the absolute difference is greater than the other hyperparameter \texttt{inter\_threshold}, \ie $B_{ij} = \{l~|~|e^l_i - \bar{e}^l_j| > \texttt{inter\_threshold}, l\in[1,k]\}$, which are the dimensions representing the attribute difference between $e_i$ and $\bar{e}_j$.} We then take the intersection across all attribute difference, $B_Y^{inter} = \bigcap_{e_i \in S_Y, \bar{e}_j \in S_{\bar{Y}}} B_{ij}$, which is the least common difference, referred to as \emph{inter-class difference} of AOI.

Finally, we take $C_Y = A_Y^{intra} \cup  B_Y^{inter}$ since the usage of $B_Y^{inter}$ is to find the representative dimensions unseen by $A_Y^{intra}$, as elaborated in~\cref{subsec:ablation_study}. A pseudo code of the method is shown in~\cref{alg:similarity}, which can be extended for all other AOIs and all \emph{protected attributes}. With \emph{intra-class similarity} and \emph{inter-class difference}, we can find the representative dimensions for both protected and facial attributes so that the generated images can properly capture the characteristics of AOI. In general, we translate the attribute appearance in images to the attribute representation in latent space. 

\begin{figure}[t]
\begin{center}
  \includegraphics[width=0.96\linewidth]{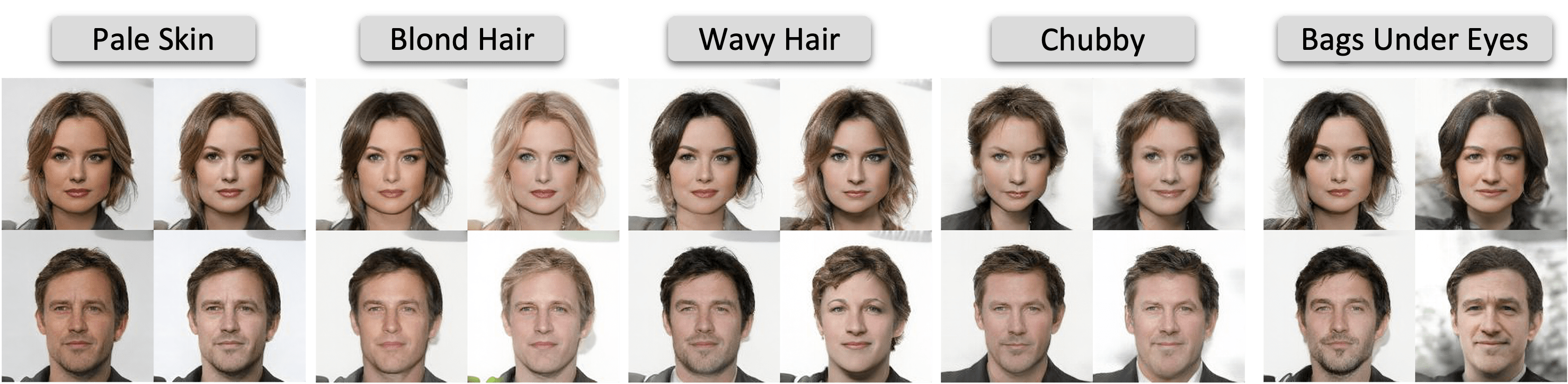}
\end{center}
  \caption{Examples of generated images with the \emph{protected attribute} and a single assigned facial attribute, negative samples (\textbf{left}) and positive samples (\textbf{right}) in each $2 \times 2$ grid.}
\label{fig:CAT}
\end{figure}

To construct the synthetic dataset $\mathcal{D}_{syn}$, we controllably translate the found attribute representations in the latent space back to image space. We first randomly simulate another set of latent vectors, referred to as \emph{identity seeds} $S_{ID} \subset \mathcal{E}$ following standard normal distribution $N(0,1)$. Then, to assign a facial attribute $Y$, we perturb $e_{ID} \in S_{ID}$ to be $e_{Y} \in S_{Y}$ for the dimension indices in $C_Y$. In parallel, the label can be automatically assigned as $Y$. In practice, the latent space consists of $R$ resolutions, and each resolution is a $k$-dimensional latent space. To further ensure the appearance of assigned facial attributes and mitigate the intervention from other facial attributes across resolution spectrum, although we do \emph{not} assume chosen dimensions are independent between different attributes and the latent space is \emph{not} disentangled by design, we subdivide the resolution spectrum to assign different kinds of facial attributes in different resolutions (\eg high-level facial attributes in lower resolutions and smaller scale facial attributes in higher resolutions), which is elaborated in~\cref{subsec:attribute_recognition}.

By generating balanced synthetic dataset $\mathcal{D}_{syn}$ with desired attributes and supplementing the minority group of the original dataset with sufficient images at both \emph{protected attribute} level and facial attribute level, we can ensure the requirements of all three fairness criteria in~\cref{DP,EO,EOS}. From the perspective of information theory, assuming the whole information contained in original dataset is fixed, \ie the best recognition performance trained on this dataset is upper-bounded, synthetic datasets generated by our method tries the best to transfer and express the recognition task-related information in a fairer way.

\noindent
\textbf{Advantages.}
By \emph{intra-class similarity} and \emph{inter-class difference}, we can accurately and effectively find a specific set of dimension indices representing AOI in the dimension spectrum so that we only need a small size of \emph{attribute seeds} with annotations as starting seeds. Beside with the resolution separation, we have a two-dimensional separation in both latent space and resolution. Thus, we can assign multiple AOIs to one single identity seed and stably preserve randomness of other unassigned facial attributes, which endows synthetic dataset to be constructed under different facial attribute distribution with more freedom. Furthermore, we assign the attribute label to the generated image $I_i = G(e_i)$ naturally with the facial attribute $\mathcal{Y}$ represented by $S_Y$ to which its latent vectors $e_i \in S_Y$ belongs, rather than rely on a shallow classifier.
% \subfile{sections/04_comparison}
\section{Experimental Evaluation}
\label{sec:experiments}
We first investigate the nature of \emph{dataset bias} by exploring each facial attribute and focus on the facial attributes which induce much bias. Furthermore, we empirically investigate the effectiveness of our method on two fashion tasks involving bias --- (1) sex classification and (2) facial attribute classification. In sex classification, we concurrently manipulate multiple \emph{femininity/masculinity attributes} to generate paired images with female and male appearance and construct sex-level balanced dataset. Meanwhile, in facial attribute classification, in the basis of sex-level balanced dataset, we further consider the other non-sex-related facial attributes to construct the semi-synthetic dataset which is balanced in terms of both sex and facial attribute. In both experiments, we first show that the classification performance of the model trained with synthetic datasets achieves comparable performance as the model trained on the original dataset. Then, we conduct a comprehensive fairness evaluation with a wide range of bias assessment metrics~\cite{directional_bias_amplification,SensitiveNets,divergence_between_score_distributions,correlation_distance,RLB} to show that fairness has been improved after training with synthetic datasets compared to the model solely trained on the original dataset. Further, we demonstrate that our method outperforms both strategic resampling~\cite{debias_resampling1} and balanced dataset construction method with synthetic images~\cite{hat_glasses_correlation} to address \emph{dataset bias}, and several debiasing models~\cite{debias_adversarial_uniform_confusion_ACC_unbalanced_dataset2_LAOFIW, Image_caption_Bias_amplification_yz,debias_domain_independent_training_BA_usage} to address \emph{task bias}. Finally, we present an ablation study to evaluate different factors influencing the performance of our method. Although we discuss sex as the \emph{protected attribute} in this section, our method is  general purpose and can be used for all \emph{protected attributes} if the labels are available.

\subsection{Attributes Study}
\label{subsec:attributes_study}
Before presenting the improvement on fairness, we first conduct in-depth evaluation of \emph{dataset bias} at the facial attribute level for CelebA dataset~\cite{CelebA}, which is a face dataset containing 202,599 images of celebrity faces and 40 binary attributes per image. Since our method produces a facial attribute-level balanced synthetic dataset, it is valuable to study \emph{dataset bias} at facial attribute level instead of the general overall accuracy for all facial attributes. Following~\cite{hat_glasses_correlation}, we train a multitask facial attribute classifier with ResNet-50~\cite{ResNet50} to recognize facial attributes on the sex-level balanced CelebA dataset. The main results are shown in~\cref{tab:multi_attributes_recognition} as baseline and full results are presented in the appendix.
 
As shown in~\cref{tab:multi_attributes_recognition} by results of baseline, solely balancing across sex does not guarantee fairness in facial attribute classification since even with balanced training across females and males, the imbalance of facial attribute across sex still exists. Although we know classification models tend to learn distinguishable representations from positive samples instead of negative samples~\cite{negative_positive}, there are insufficient positive samples of some specific facial attributes in the minority group. Inspired by the categorization in~\cite{hat_glasses_correlation} \jiazhi{where they only categorize a subset of facial attributes,} we summarize all 40 facial attributes in CelebA dataset into three groups --- (1) \emph{unbiased attributes} which do not yield much bias (\ie the difference of AP between female and male is less than 5\%), (2) \emph{masculinity/femininity attributes}, which are considered as AOI in~\cref{subsec:sex_classfication} to construct sex-level balanced dataset, and (3) non-sex-related facial attributes but inducing much bias even with sex-level balanced training, which are considered as AOI appending \emph{masculinity/femininity attributes} in~\cref{subsec:attribute_recognition} to construct both sex-level and facial attribute-level balanced dataset.

\begin{table}[t]
\centering
\tiny
\caption{Performance and fairness comparison on sex classification.}
\label{tab:sex_classification}
\resizebox{1\textwidth}{!}{%
\begin{tabular}{cc ccc ccc ccc}
\toprule
                                           &            & \multicolumn{3}{c}{Number of Training Images} & \multicolumn{3}{c}{Classification Accuracy $\uparrow$} & Information Leakage & \multicolumn{2}{c}{Statistical Dependence} \\ 
                                           \cmidrule(r){3-5} \cmidrule(){6-8} \cmidrule(lr){9-9} \cmidrule(){10-11}
                                           &            & Female         & Male           & Total         & Female        & Male          & Overall       & BA $\downarrow$                 & ${dcor}^2 \downarrow$                 & RLB $\downarrow$                \\ \midrule
\multirow{2}{*}{Origin training   dataset} & Imbalanced & 94509          & 68261          & 162770        & \textbf{99.2} & 97.7          & \textbf{98.6} & -0.015              & 0.656                & 3.854               \\
                                           & Balanced   & 68261          & 68261          & 136522        & \textbf{99.2} & 97.6          & 98.5          & -0.016              & 0.651                & 3.792               \\
\multirow{3}{*}{GAN-Debiasing~\cite{hat_glasses_correlation}}             & Original     & 254509         & 228261         & 482770        & 99.1          & 97.4          & 98.4          & \textbf{-0.018}     & 0.596                & 3.567               \\
                                           & Same size  & 68261          & 68261          & 136522        & 98.8          & 97.2          & 98.0          & -0.017              & 0.583                & 3.578               \\
                                           & Supplement & 94509          & 94509          & 189018        & \textbf{99.2} & 97.2          & 98.2          & -0.016              & 0.574                & 3.574               \\
\multirow{2}{*}{Ours}                      & Same size  & 68261          & 68261          & 136522        & 99.1          & 97.3          & 98.3          & \textbf{-0.018}     & 0.592                & 3.742               \\
                                           & Supplement & 94509          & 94509          & 189018        & \textbf{99.2} & \textbf{97.8} & 98.5          & -0.016              & \textbf{0.544}       & \textbf{3.481}     \\ \bottomrule
\end{tabular}%
}
\end{table}

\subsection{Synthetic Attribute-level Balanced Datasets}
\label{subsec:synthetic_dataset}
To construct the synthetic dataset $\mathcal{D}_{syn}$, we train the generative model solely on the training set of CelebA dataset containing 162,770 images to ensure the isolation from the testing set since the images generated by trained GAN are used for the following tasks. For the generative model choice to generate high-quality images (which is elaborated in appendix), we use StyleGAN2~\cite{styleGan2} since \emph{style mixing regularization} in StyleGAN2 facilitates our goal to assign different specific facial attributes and mitigate the interference from other unassigned facial attributes. We persist in training the generative model to generate more real images. As a standard metric, FID is used to evaluate sample qualities of generated images. FID of the presented results in the paper is $3.56$, which is comparable to existing GANs~\cite{FID_comparable1, FID_comparable2} trained on CelebA dataset and posted recently. Although the benchmark for FID of image generation on CelebA dataset is $2.71$~\cite{FID_benchmark_CelebA_64}, our objective is to utilize generated images to improve fairness of existing datasets rather than improve image quality of generated images.

\begin{table}[t]
\centering
\caption{Performance and fairness comparison on facial attribute classification.}
\label{tab:multi_attributes_recognition}
\resizebox{1\textwidth}{!}{%
\begin{tabular}{cccccccccccc}
\toprule
                      &               & Chubby        & BigNose        & WavyHair       & PaleSkin       & BlondHair      & DoubleChin     & PointyNose     & BagsUnderEyes  & HighCheekbones & Average        \\
                      \cmidrule(lr){3-3} \cmidrule(lr){4-4} \cmidrule(lr){5-5} \cmidrule(lr){6-6} \cmidrule(lr){7-7} \cmidrule(lr){8-8} \cmidrule(lr){9-9} \cmidrule(lr){10-10} \cmidrule(lr){11-11} \cmidrule(lr){12-12}
\multirow{4}{*}{AP $\uparrow$} & Baseline      & \textbf{62.1} & \textbf{71.2}  & \textbf{88.2}  & \textbf{69.1}  & 92.0           & \textbf{61.2}  & \textbf{64.9}  & \textbf{67.8}  & 95.4           & \textbf{74.7}  \\
     & Resampling~\cite{debias_resampling1}    & 54.5          & 62.2           & 87.2           & 67.2           & 87.0           & 46.4           & 61.7           & 58.3           & 95.1           & 68.8           \\
     & GAN-Debiasing~\cite{hat_glasses_correlation} & 58.8          & 69.5           & 85.6           & 68.5           & 90.6           & 58.5           & 63.5           & 63.1           & 95.3           & 72.6           \\
     & Ours          & 60.5          & 66.5           & 86.8           & 69.0           & \textbf{92.1}  & 58.6           & 61.4           & 62.9           & \textbf{95.5}  & 72.6           \\ \midrule
\multirow{4}{*}{DEO $\downarrow$} & Baseline      & 28.7          & 35.6           & 33.0           & 11.9           & 10.8           & 26.7           & 32.8           & 20.1           & 10.8           & 23.4           \\
     & Resampling~\cite{debias_resampling1}    & \textbf{15.0} & 12.7           & 12.1           & \textbf{6.2}   & 4.3            & 4.2            & 5.2            & 19.2           & 5.9            & 9.4            \\
     & GAN-Debiasing~\cite{hat_glasses_correlation} & 26.6          & 32.7           & 23.3           & 10.8           & 3.6            & 22.1           & 28.8           & 16.8           & 9.3            & 19.3           \\
     & Ours          & 21.2          & \textbf{3.9}   & \textbf{11.8}  & 9.3            & \textbf{3.5}   & \textbf{3.3}   & \textbf{3.1}   & \textbf{15.0}  & \textbf{4.3}   & \textbf{8.4}   \\ \midrule
\multirow{4}{*}{BA $\downarrow$} & Baseline      & 1.72          & 5.83           & -3.27          & 1.19           & 0.12           & 0.46           & 4.78           & 2.27           & -1.32          & 1.31           \\
     & Resampling~\cite{debias_resampling1}    & 1.38          & \textbf{-8.69} & \textbf{-5.49} & -0.30          & -3.37          & \textbf{-1.73} & -3.21          & \textbf{-9.44} & \textbf{-6.17} & \textbf{-4.11} \\
     & GAN-Debiasing~\cite{hat_glasses_correlation} & 1.30          & 5.51           & -3.60          & 0.20           & 0.50           & 0.46           & 3.71           & 1.14           & -1.89          & 0.81           \\
     & Ours          & \textbf{1.16} & -6.71          & -5.10          & \textbf{-0.44} & \textbf{-4.61} & -1.16          & \textbf{-5.35} & -5.95          & -4.27          & -3.60          \\ \midrule
\multirow{4}{*}{KL $\downarrow$} & Baseline      & 0.39          & 0.60           & 0.42           & 0.08           & 0.46           & 0.30           & 0.32           & 0.27           & 0.12           & 0.33           \\
     & Resampling~\cite{debias_resampling1}    & \textbf{0.18} & \textbf{0.13}  & 0.15           & 0.07           & \textbf{0.05}  & 0.17           & 0.21           & 0.26           & 0.12           & 0.15           \\
     & GAN-Debiasing~\cite{hat_glasses_correlation} & 0.37          & 0.54           & 0.33           & 0.07           & 0.37           & 0.29           & 0.20           & 0.26           & 0.09           & 0.28           \\
     & Ours          & 0.33          & 0.13           & \textbf{0.06}  & \textbf{0.04}  & 0.34           & \textbf{0.09}  & \textbf{0.04}  & \textbf{0.08}  & \textbf{0.08}  & \textbf{0.13}  \\ \midrule
\multirow{4}{*}{${dcor}^2 \downarrow$} & Baseline      & 0.58          & 0.65           & 0.82           & 0.53           & 0.34           & 0.58           & 0.85           & 0.86           & 0.82           & 0.67           \\
     & Resampling~\cite{debias_resampling1}    & 0.58          & 0.59           & 0.77           & 0.47           & 0.69           & 0.53           & 0.83           & 0.65           & 0.56           & 0.63           \\
     & GAN-Debiasing~\cite{hat_glasses_correlation} & 0.35          & 0.54           & 0.50           & 0.08           & \textbf{0.17}  & 0.31           & 0.39           & 0.43           & 0.33           & 0.34           \\
     & Ours          & \textbf{0.18} & \textbf{0.31}  & \textbf{0.33}  & \textbf{0.06}  & 0.23           & \textbf{0.17}  & \textbf{0.24}  & \textbf{0.25}  & \textbf{0.30}  & \textbf{0.23}  \\ \midrule
\multirow{4}{*}{RLB $\downarrow$} & Baseline      & 0.63          & 1.24           & 1.56           & 0.03           & 1.12           & 0.45           & 0.55           & 0.82           & 1.36           & 0.86           \\
     & Resampling~\cite{debias_resampling1}    & 0.58          & 1.13           & 1.45           & 0.03           & 1.03           & 0.44           & 0.50           & 0.63           & 0.96           & 0.75           \\
     & GAN-Debiasing~\cite{hat_glasses_correlation} & 0.34          & 1.14           & 1.42           & \textbf{0.01}  & 0.86           & 0.31           & 0.49           & 0.61           & 0.69           & 0.65           \\
     & Ours          & \textbf{0.28} & \textbf{1.11}  & \textbf{1.30}  & \textbf{0.01}  & \textbf{0.73}  & \textbf{0.30}  & \textbf{0.09}  & \textbf{0.43}  & \textbf{0.39}  & \textbf{0.52} \\ \bottomrule
\end{tabular}%
}
\end{table}

\begin{figure}[t]
\begin{center}
  \includegraphics[width=0.5\linewidth]{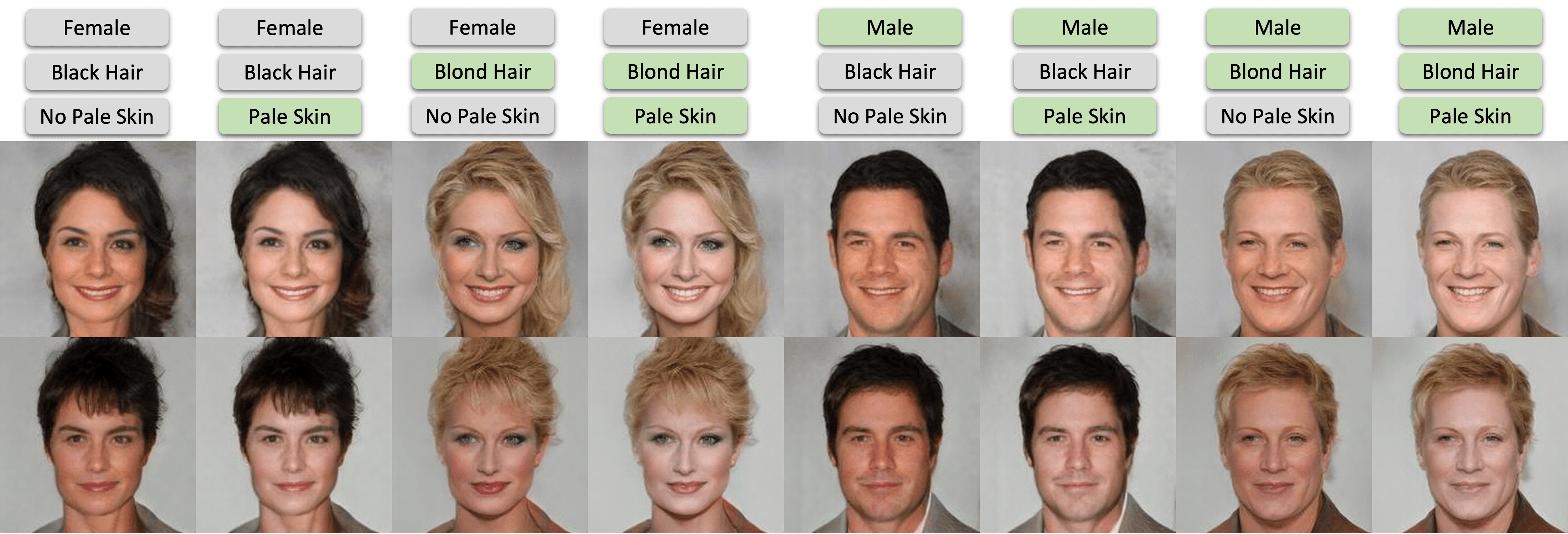}
\end{center}
  \caption{Examples of generated images with multiple assigned facial attributes.}
\label{fig:grid}
\end{figure}

To assign attributes, we use $\texttt{intra\_threshold}  = 2\sqrt{2}$ and $\texttt{inter\_threshold}$ 
$= \sqrt{2}$. In general, we empirically found that any settings for these two hyperparameters among the recommended range $[\sqrt{2}, 2\sqrt{2}]$ will not significantly affect the performance of downstream classification tasks, which confirms with our theoretical discussion in~\cref{subsec:ablation_study}. Furthermore, for resolution spectrum separation, StyleGAN2~\cite{styleGan2} have already provided reference to assign different attributes in different resolutions, \eg high-level attributes (hair style, face shape) in coarse spatial resolution $(4^2 - 8^2)$, and racial appearance (colors of eyes, hair, skin) in fine spatial resolution $(16^2 - 1024^2)$. The dimensionality of the latent space is 512 and the resolution of generated images is set to be $256^2$ so that by~\cite{styleGan2} we generally have a 14-layer 512-dimensional latent space to assign attributes. Specifically, we leave $4^2$ original for the basic identity construction. For resolution choices to assign attributes, we assign face shape attributes (\textsf{Chubby}, \textsf{Big Nose}, \textsf{Pointy Nose}, \textsf{High Cheekbones} and \textsf{Double Chin}) in $8^2$, fine face shape attributes (\textsf{Bags Under Eyes}, \textsf{Wavy Hair} and \textsf{Straight Hair}) in $16^2$, hair color attributes (\textsf{Black Hair}, \textsf{Blond Hair}, \textsf{Brown Hair} and \textsf{Gray Hair}) in $32^2 - 64^2$, and skin color attributes (\textsf{Pale Skin}) in $128^2 - 256^2$. Besides, we assign \emph{masculinity/femininity attributes} in $8^2 - 16^2$. As shown in~\cref{fig:sex}, the generated images are listed in pairs with highlighted \emph{femininity/masculinity attributes}, and similar non-sex facial attributes (\eg skin color and hair color) and other photo settings (\eg lighting, pose and background). Besides, in~\cref{fig:CAT}, we controllably assign \emph{femininity/masculinity attributes} and another facial attribute. Furthermore, in~\cref{fig:grid}, we concurrently assign sex and multiple facial attributes and generally keep other unassigned facial attributes unchanged, which demonstrates the advantage that attribute assignment does not interfere with each other.

\subsection{Sex Classification}
\label{subsec:sex_classfication}
As in~\cite{FairFace,PPB_metric_Joy}, the imbalance of training dataset may hamper the accuracy of sex classification, we therefore validate the fairness improvement of our synthetic datasets on sex classification in this section. To help downstream classification models to well learn the difference between female and male (the useful information for sex classification), given an identity seed, we only modify \emph{femininity/masculinity attributes} and leave other facial attributes unchanged, as shown in~\cref{fig:sex}, which also ensures that there are sufficient samples with large diversity of facial attributes from identity seeds in both female group and male group.

\begin{table}[t]
\centering
\scriptsize
\caption{Comparison of debiasing models on facial attribute classification.}
\label{tab:comparison}
\resizebox{\textwidth}{!}{%
\begin{tabularx}{\textwidth}{l *{6}{Y} cc}
\toprule
                               & \multicolumn{3}{c}{mAP $\uparrow$}                       & \multicolumn{3}{c}{Information Leakage $\downarrow$}       & \multicolumn{2}{c}{Statistical Dependence $\downarrow$} \\ \cmidrule(){2-4} \cmidrule(lr){5-7} \cmidrule(){8-9}
                               & Female        & Male          & Overall       & DEO          & BA             & KL            & ${dcor}^2$~\cite{correlation_distance}                 & RLB~\cite{RLB}                \\ \midrule
Baseline                       & \textbf{75.3} & 72.2          & 74.1          & 20.9         & 0.99           & 0.27          & 0.85                 & 1.33                \\ 
Adversarial forgetting~\cite{debias_adversarial_uniform_confusion_ACC_unbalanced_dataset2_LAOFIW} & 73.2          & 70.4          & 72.1          & 18.8         & 0.61           & 0.25          & 0.51                 & 1.25                \\ 
Domain adaptation~\cite{Image_caption_Bias_amplification_yz}              & 74.5          & 72.3          & 73.7          & 19.7         & 0.82           & 0.21          & 0.48                 & 1.28                \\ 
Domain independent~\cite{debias_domain_independent_training_BA_usage}             & 74.6          & \textbf{74.2} & \textbf{74.4} & 20.8         & 0.55           & 0.27          & 0.37                 & 1.12                \\ 
Ours                           & 73.0          & 73.2          & 73.1          & \textbf{8.4} & \textbf{-3.26} & \textbf{0.16} & \textbf{0.24}        & \textbf{0.65}       \\ \bottomrule
\end{tabularx}%
}
\end{table}

\noindent
\textbf{Experiment setup.}
ResNet-50~\cite{ResNet50} is used as the backbone for sex classifier. The baseline model is trained with original CelebA dataset~\cite{CelebA} under two settings --- (1) \texttt{Original} to keep the training set of CelebA dataset fully original with the whole set of 162,770 images, and (2) \texttt{Balanced} to balance the training set across sex. For training with synthetic dataset, we also conduct two types of experiments --- (1) \texttt{Same size} to combine one half of balanced original training set with a same size balanced synthetic dataset to be mixed dataset so that the number of mixed dataset is the same as the number of training images in the balanced original dataset, and (2) \texttt{Supplement} to supplement the lack of male images compared with female images in the original dataset so that the mixed dataset is balanced in the whole dataset scale. For comparison with the other attribute-level balance method~\cite{hat_glasses_correlation}, we conduct experiments under the original setting in their paper (the original CelebA dataset combined with the balanced synthetic dataset with 160,000 sex pairs of images) and two same settings of ours. The testing set is the original testing CelebA dataset for all experiments.

\noindent
\textbf{Evaluation protocol.}
To demonstrate the effectiveness of our method on sex classification and facial attribute classification, we use accuracy and fairness as the main evaluation metrics. First, we compare the classification performance between the model trained on synthetic datasets and the original dataset to verify that the testing accuracy on the real testing dataset is preserved at the same level, which demonstrates that the generated images yield proper facial attributes in the metrics of the real dataset. Meanwhile, we use several fairness metrics for a comprehensive fairness comparison since different types of metrics assess bias in different directions. According to the taxonomy of bias assessment metrics in~\cite{RLB}, we select several representative metrics in these categories. We select Bias Amplification (BA)~\cite{directional_bias_amplification,debias_domain_independent_training_BA_usage} among \emph{information leakage}-based metrics and two types of metrics based on \emph{statistical dependence}, Distance Correlation (${dcor}^2$)~\cite{correlation_distance, BR_net_correlation_distance_bias_metric} and Representation-Level Bias (RLB)~\cite{RLB} to increase the diversity of metrics and comprehensively evaluate fairness in different ways.

\noindent
\textbf{Results.}
As shown in~\cref{tab:sex_classification}, we can see the testing accuracy of synthetic dataset is kept at the same level under both two settings, which demonstrates that the generated images yield proper \emph{femininity/masculinity attributes} in the original dataset. Furthermore, although the classification accuracy is already saturated, with additional images under the \texttt{Supplement} setting, we improve the accuracy in minority groups. On the other hand, for fairness evaluation compared with other datasets, our method further improves fairness. Although \cite{hat_glasses_correlation} uses more images than the original dataset, our method yields even better results in both accuracy and fairness metrics with fewer training images. We elaborate the discussion on the relation between the size of the synthetic dataset and performance of our method in the appendix.

\subsection{Facial Attribute Classification}
\label{subsec:attribute_recognition}
Having found that the remaining \emph{dataset bias} of facial attribute classification on CelebA dataset stems from imbalance at both \emph{protected attribute} level and facial attribute level as discussed in~\cref{subsec:attributes_study}, we construct the synthetic dataset which is balanced in both these two levels to study the effectiveness of our method for all non-sex-related facial attributes, and compare our method with methods based on sex-level balance~\cite{debias_resampling1} and facial attribute-level balance~\cite{hat_glasses_correlation}. Furthermore, we highlight the comparisons with debiasing models~\cite{debias_adversarial_uniform_confusion_ACC_unbalanced_dataset2_LAOFIW,Image_caption_Bias_amplification_yz,debias_domain_independent_training_BA_usage}.

\noindent
\textbf{Experiment Setup.}
We train a multitask facial attribute classifier with ResNet-50~\cite{ResNet50}. The baseline is trained with sex-level balanced CelebA dataset constructed from original CelebA dataset. Our method combines a portion of original dataset without surplus samples of some facial attributes and the synthetic dataset which supplements insufficient samples of AOI so that the mixed dataset is balanced at both sex level and facial attribute level, as illustrated in~\cref{fig:idea}.

\noindent
\textbf{Evaluation protocol.}
We use Average Precision (AP) as the performance metric. For bias assessment, beside BA, ${dcor}^2$ and RLB, Difference in Equal Opportunity (DEO)~\cite{SensitiveNets, HSIC} as a metric version of \emph{equal opportunity} and KL-divergence between score distribution (KL)~\cite{divergence_between_score_distributions,hat_glasses_correlation} as a stronger notion stemming from \emph{equalized odds} are used to verify the achievements of~\cref{EO,EOS}.

\noindent
\textbf{Results.}
In~\cref{tab:multi_attributes_recognition}, we present one representative facial attribute for the mutually exclusive facial attributes, \eg \textsf{Wavy Hair} and \textsf{Straight Hair}, and full results are presented in appendix. As AP is preserved at the same level for all AOIs, our method outperforms other methods for \textsf{Blond Hair} and \textsf{High Cheekbones}. In fairness comparison with baseline, all three methods introduce fairness under all metrics in a same trend. As pointed by~\cite{RLB}, ${dcor}^2$ and RLB, the metrics based on \emph{statistical dependence} are more consistent and stable. In this sense, although strategically resampling~\cite{debias_resampling_representation_bias_colored_MNIST} outperforms our method under BA, it yields second worse performance on the metrics based on \emph{statistical dependence}. Furthermore, with better DEO and KL, we verify the achievement of~\cref{EO,EOS} in~\cref{sec:method}, which is unreachable for sex-level balanced dataset and strategically resampling. According to results of imbalanced training, due to the existence of the facial attribute-level skew in the dataset, the classifier trained with these skew datasets may be biased already. In this sense, compared with~\cite{hat_glasses_correlation}, our method is more stable in all facial attributes since we assign label directly rather than rely on such unreliable facial attribute classifier trained on original dataset as in~\cite{hat_glasses_correlation}, which may be harmful for fairness. Compared with~\cref{tab:sex_classification}, the range of two metrics based on \emph{statistical dependence} in sex classification are clearly higher than the range in the facial attribute classification since they directly reveal the statistical dependence between learned representations to predict sex and sex labels themselves, which better assess the bias for sex classification. 

We also conduct an overall comparison of average results over all non-sex-related AOIs (7 facial attributes in the paper and 7 mutually exclusive facial attributes in appendix) with several debiasing models based on adversarial forgetting~\cite{debias_adversarial_uniform_confusion_ACC_unbalanced_dataset2_LAOFIW}, domain adaptation~\cite{Image_caption_Bias_amplification_yz} and domain independent~\cite{debias_domain_independent_training_BA_usage}. We train a ResNet-50~\cite{ResNet50} to classify all non-sex-related AOIs on the original CelebA dataset as baseline model. For the multi-label classification, mean Average Precision (mAP) is used to evaluate the overall performance. In~\cref{tab:comparison} of classification performance comparison, domain independent method outperforms other methods. In parallel, without the in-process debiasing techniques usage, our method yields comparable classification performance and outperforms other methods in all bias assessment metrics, which demonstrates a comprehensive fairness improvement. In general, there is a tradeoff between classification performance and fairness performance, \ie although domain independent method outperforms other methods in mAP, our method improves fairness mostly at an acceptable expense of classification performance. Furthermore, the proposed synthetic dataset is model-agnostic as compared under different backbones in appendix.

\subsection{Ablation Study}
\label{subsec:ablation_study}

\begin{figure}[t]
\centering
\begin{minipage}[b]{.5\linewidth}
    \centering
    \subfloat[][{Deprecated.}]{\label{fig:deprecated}\includegraphics[width=.78\linewidth]{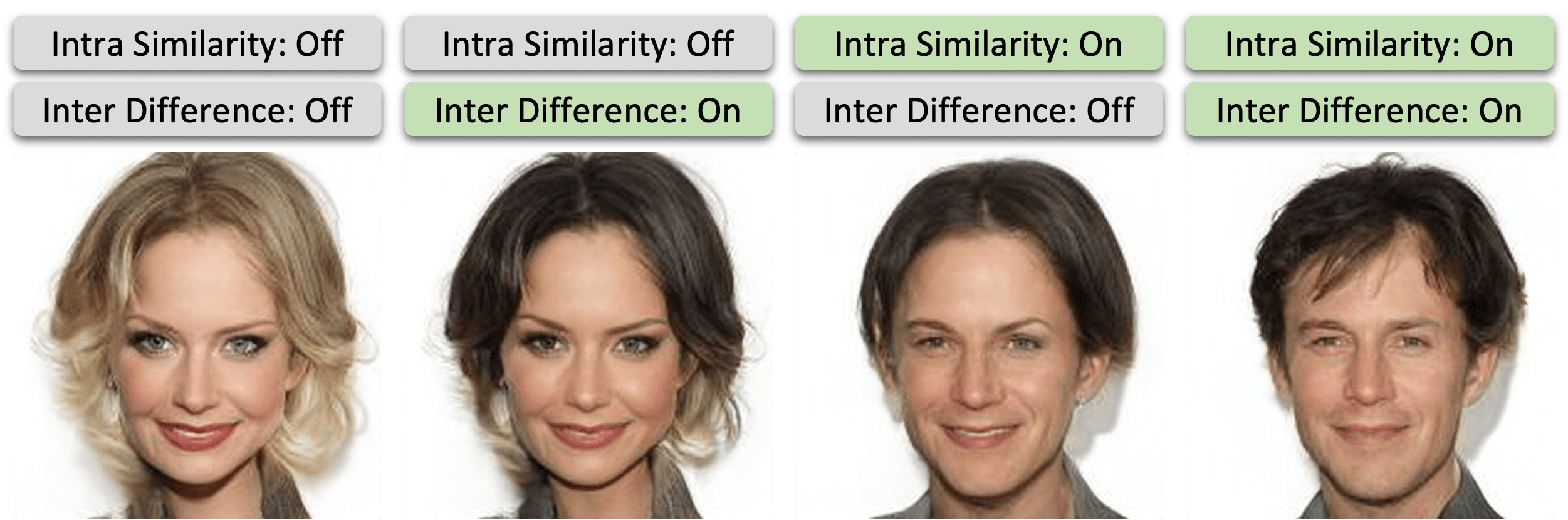}}
\end{minipage}
\begin{minipage}[b]{.4\linewidth}    
  \centering
    \subfloat[][Overemphasized.]{\label{fig:overemphasized}\includegraphics[width=.6\linewidth]{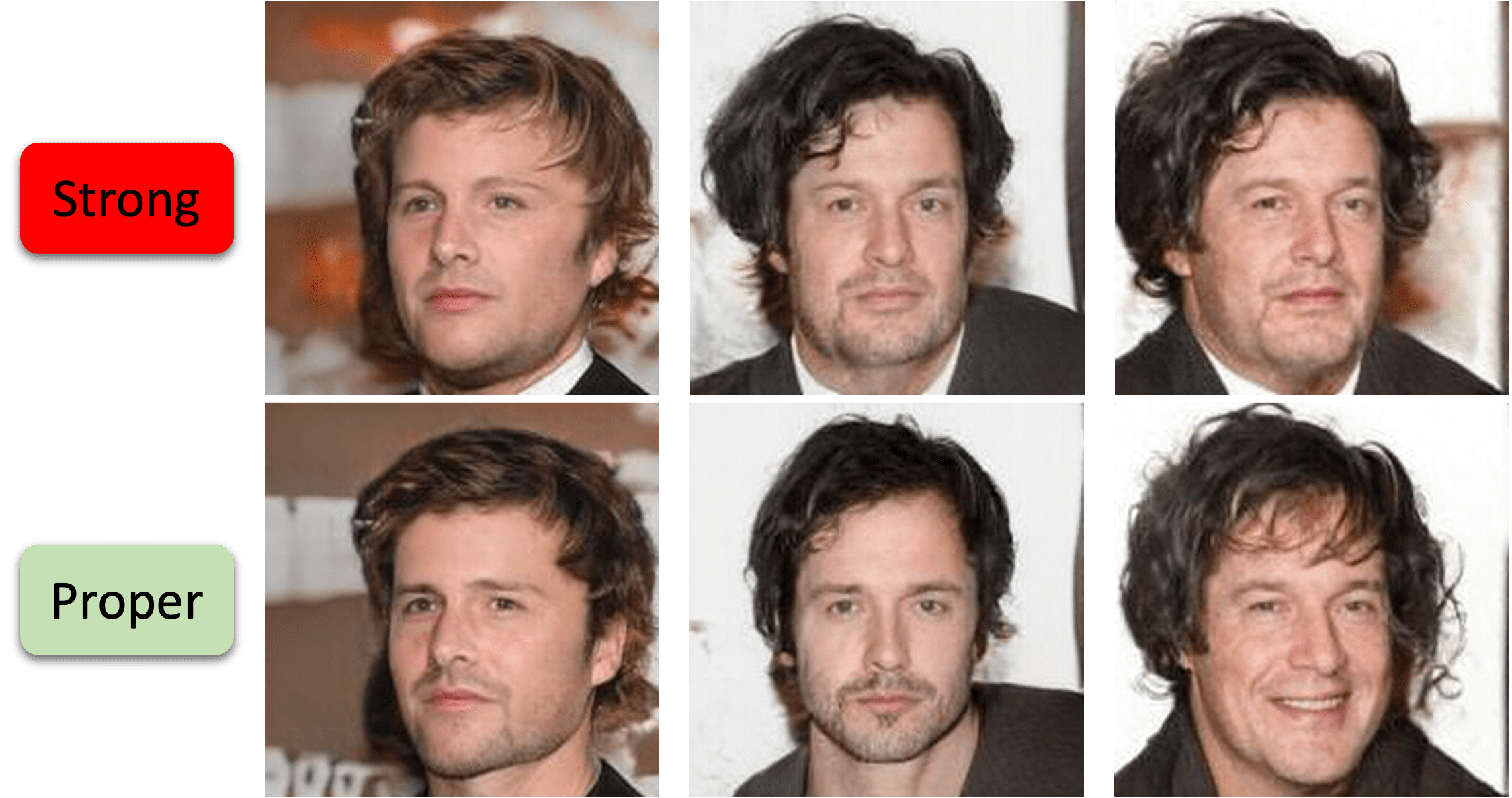}}
\end{minipage}
    \caption{The role of intra-class similarity and inter-class difference.}
\label{fig:intra_inter}
\end{figure}

\noindent
\textbf{\emph{Intra-class similarity} and \emph{inter-class difference}.}
In this section, we will discuss \emph{intra-class similarity} and \emph{inter-class difference} involving two hyperparameter \texttt{intra\_threshold} and \texttt{inter\_threshold}. Considered the task to generate female and male images for which, given \emph{masculinity attributes seeds} $S_{male}$ and \emph{femininity attributes seeds} $S_{female}$, \emph{intra-class similarity} is deprecated when \texttt{intra\_threshold} is small since there may be no dimension indice where $\forall e_i, e_j \in S_{male}, |e_i - e_j| <$ \texttt{intra\_threshold}, \ie $A^{intra}_{male}$ is empty. By contrast, \emph{inter-class difference} will be deprecated when $inter\_threshold$ is large. For example, given an identity latent vector $e_{ID}$ whose generated image is female, when \emph{intra-class similarity} and \emph{inter-class difference} are both deprecated, the generated image is same as the prime image generated from $e_{ID}$. As shown in~\cref{fig:deprecated}, without \emph{intra-class similarity}, the generated image is apparently female. Further, when we apply \emph{intra-class similarity} and deprecate \emph{inter-class difference}, the generated image is transferred to be male with few \emph{femininity attributes} (\eg hair) since the generated model has learned \emph{masculinity attributes} but has not fully suppressed \emph{femininity attributes}. On the other hand, if \texttt{intra\_threshold} is large or \texttt{inter\_threshold} is small, \ie $A^{intra}_{male}$ or $B^{inter}_{male}$ contains all dimension indices so that \emph{intra-class similarity} or \emph{inter-class difference} is overemphasized. As shown in~\cref{fig:overemphasized}, compared with generated images under proper threshold settings, the generated images under strong threshold settings in the top row are cookie-cutter without the randomness from identity seeds. In the whole process, \emph{intra-class similarity} contributes mostly and \emph{inter-class difference} plays an auxiliary role. Since the random latent vector is simulated under standard normal distribution $N(0,1)$ and the difference between two latent vectors follows normal distribution $N(0,\sqrt{2})$ where standard deviation is $\sqrt{2}$, the proper range for two thresholds is $[\sqrt{2},2\sqrt{2}]$ so that $A_Y^{inter}$ may contain the dimension indices in the main lope representing attribute similarity among $S_Y$, and $B_Y^{intra}$ may exclude the dimension indices in the main lope representing attribute similarity of $S_Y$ and $S_{\bar{Y}}$, where $S_{\bar{Y}}$ is the set of latent vectors without \emph{Attribute of Interest} $\mathcal{Y}$.

\section{Conclusion}
By the investigation on the imbalance at the facial attribute level, we can clearly find the root cause for the uneven performance across \emph{protected attributes}. To address the imbalance, from the perspective of information theory, we try our best to mitigate the loss of information from original data (\ie comparable performance) and orderly express such information in a fairer way (\ie improved fairness) while the amount of whole information is fixed. In comparison with debiasing models, we have found fairness-performance tradeoff in model training, as fairness-efficiency tradeoff in the real world. Still, the valuable remark is that although the network trained with the proposed synthetic data does not outperform the debiasing models in recognition performance, it is impressive that synthetic images can achieve consistent performance with real data and further yields better fairness. Furthermore, in the image generation process, our method overcomes the difficulty to obtain labels by \emph{intra-class similarity} and \emph{inter-class difference} instead of relying on a shallow attribute classifier or significant extra human annotations. Therefore, our method can be scalably used in the model fully trained with synthetic datasets in the future. Besides, it is a good direction for future researchers to extend our method to other face datasets and introduce fairness for more facial attributes. Finally, our work presents a good example for the probability of the same-level performance for computer vision models partially or fully trained with synthetic dataset.

\noindent
\textbf{Acknowledgement}: This research is based upon work supported in part by the Office of the Director of National Intelligence (ODNI), Intelligence Advanced Research Projects Activity (IARPA), via [2022-21102100007]. The views and conclusions contained herein are those of the authors and should not be interpreted as necessarily representing the official policies, either expressed or implied, of ODNI, IARPA, or the U.S. Government. The U.S. Government is authorized to reproduce and distribute reprints for governmental purposes notwithstanding any copyright annotation therein.

\clearpage
{
\bibliographystyle{splncs04}
\bibliography{reference}

\begin{thebibliography}{10}
\providecommand{\url}[1]{\texttt{#1}}
\providecommand{\urlprefix}{URL }
\providecommand{\doi}[1]{https://doi.org/#1}

\bibitem{def_dataset_bias}
Adeli, E., Zhao, Q., Pfefferbaum, A., Sullivan, E.V., Fei-Fei, L., Niebles,
  J.C., Pohl, K.M.: Representation learning with statistical independence to
  mitigate bias. In: Proceedings of the IEEE/CVF Winter Conference on
  Applications of Computer Vision. pp. 2513--2523 (2021)

\bibitem{BR_net_correlation_distance_bias_metric}
Adeli, E., Zhao, Q., Pfefferbaum, A., Sullivan, E.V., Fei-Fei, L., Niebles,
  J.C., Pohl, K.M.: Representation learning with statistical independence to
  mitigate bias. In: Proc. of the IEEE/CVF Winter Conf. on Applications of
  Computer Vision. pp. 2513--2523 (2021)

\bibitem{GANs_Objectives}
Alaa, A.M., van Breugel, B., Saveliev, E., van~der Schaar, M.: How faithful is
  your synthetic data? sample-level metrics for evaluating and auditing
  generative models. arXiv preprint arXiv:2102.08921  (2021)

\bibitem{debias_adversarial_uniform_confusion_ACC_unbalanced_dataset2_LAOFIW}
Alvi, M., Zisserman, A., Nell{\aa}ker, C.: Turning a blind eye: Explicit
  removal of biases and variation from deep neural network embeddings. In:
  Proceedings of the European Conference on Computer Vision (ECCV) Workshops.
  pp.~0--0 (2018)

\bibitem{transect}
Balakrishnan, G., Xiong, Y., Xia, W., Perona, P.: Towards causal benchmarking
  of biasin face analysis algorithms. In: Deep Learning-Based Face Analytics,
  pp. 327--359. Springer (2021)

\bibitem{AI_fairness_360}
Bellamy, R.K., Dey, K., Hind, M., Hoffman, S.C., Houde, S., Kannan, K., Lohia,
  P., Martino, J., Mehta, S., Mojsilovic, A., et~al.: Ai fairness 360: An
  extensible toolkit for detecting, understanding, and mitigating unwanted
  algorithmic bias. arXiv preprint arXiv:1810.01943  (2018)

\bibitem{debias_resampling1}
Bickel, S., Br{\"u}ckner, M., Scheffer, T.: Discriminative learning under
  covariate shift. J. of Machine Learning Research  \textbf{10}(9) (2009)

\bibitem{sex_race_PPB}
Buolamwini, J., Gebru, T.: Gender shades: Intersectional accuracy disparities
  in commercial gender classification. In: Friedler, S.A., Wilson, C. (eds.)
  Proc. of the 1st Conf. on Fairness, Accountability and Transparency. Proc. of
  Machine Learning Research, vol.~81, pp. 77--91. PMLR (23--24 Feb 2018),
  \url{http://proceedings.mlr.press/v81/buolamwini18a.html}

\bibitem{PPB_metric_Joy}
Buolamwini, J., Raji, I.D.: Actionable auditing: Investigating the impact of
  publicly naming biased performance results of commercial ai products.
  Conference on Artificial Intelligence, Ethics, and Society (2019)

\bibitem{preprocessing2}
Calmon, F.P., Wei, D., Vinzamuri, B., Ramamurthy, K.N., Varshney, K.R.:
  Optimized pre-processing for discrimination prevention. In: Proceedings of
  the 31st International Conference on Neural Information Processing Systems.
  pp. 3995--4004 (2017)

\bibitem{gan_development3}
Chan, E.R., Monteiro, M., Kellnhofer, P., Wu, J., Wetzstein, G.: pi-gan:
  Periodic implicit generative adversarial networks for 3d-aware image
  synthesis. In: Proceedings of the IEEE/CVF conference on computer vision and
  pattern recognition. pp. 5799--5809 (2021)

\bibitem{divergence_between_score_distributions}
Chen, M., Wu, M.: Towards threshold invariant fair classification. In:
  Conference on Uncertainty in Artificial Intelligence. pp. 560--569. PMLR
  (2020)

\bibitem{infogan}
Chen, X., Duan, Y., Houthooft, R., Schulman, J., Sutskever, I., Abbeel, P.:
  Infogan: Interpretable representation learning by information maximizing
  generative adversarial nets. Advances in neural information processing
  systems  \textbf{29} (2016)

\bibitem{generate_fair1}
Choi, K., Grover, A., Singh, T., Shu, R., Ermon, S.: Fair generative modeling
  via weak supervision. In: International Conference on Machine Learning. pp.
  1887--1898. PMLR (2020)

\bibitem{debias_multi_task}
Das, A., Dantcheva, A., Bremond, F.: Mitigating bias in gender, age and
  ethnicity classification: a multi-task convolution neural network approach.
  In: Proceedings of the European Conference on Computer Vision (ECCV)
  Workshops. pp.~0--0 (2018)

\bibitem{debias1_no_bam_DP}
Dwork, C., Hardt, M., Pitassi, T., Reingold, O., Zemel, R.: Fairness through
  awareness. In: Proc. of the 3rd innovations in theoretical computer science
  conf. pp. 214--226 (2012)

\bibitem{debface_representation_disentanglement}
Gong, S., Liu, X., Jain, A.K.: Jointly de-biasing face recognition and
  demographic attribute estimation. In: European Conf. on Computer Vision. pp.
  330--347. Springer (2020)

\bibitem{GAC}
Gong, S., Liu, X., Jain, A.K.: Mitigating face recognition bias via group
  adaptive classifier. In: Proc. of the IEEE/CVF Conf. on Computer Vision and
  Pattern Recognition. pp. 3414--3424 (2021)

\bibitem{GAN}
Goodfellow, I., Pouget-Abadie, J., Mirza, M., Xu, B., Warde-Farley, D., Ozair,
  S., Courville, A., Bengio, Y.: Generative adversarial nets. Advances in
  neural information processing systems  \textbf{27} (2014)

\bibitem{FRVT3}
Grother, P., Ngan, M., Hanaoka, K.: Face recognition vendor test (fvrt): Part
  3, demographic effects. National Institute of Standards and Technology (2019)

\bibitem{alignment_good1}
G{\"u}nther, M., Rozsa, A., Boult, T.E.: Affact: Alignment-free facial
  attribute classification technique. In: 2017 IEEE International Joint
  Conference on Biometrics (IJCB). pp. 90--99. IEEE (2017)

\bibitem{DP}
Hardt, M., Price, E., Srebro, N.: Equality of opportunity in supervised
  learning. In: Proc. of the 30th International Conf. on Neural Information
  Processing Systems. p. 3323–3331. NIPS'16, Curran Associates Inc., Red
  Hook, NY, USA (2016)

\bibitem{gan_augment_real_data1_low_shot}
Hariharan, B., Girshick, R.: Low-shot visual recognition by shrinking and
  hallucinating features. In: Proceedings of the IEEE International Conference
  on Computer Vision. pp. 3018--3027 (2017)

\bibitem{ResNet50}
He, K., Zhang, X., Ren, S., Sun, J.: Deep residual learning for image
  recognition. In: Proc. of the IEEE conf. on computer vision and pattern
  recognition. pp. 770--778 (2016)

\bibitem{synthetic_data_face_recognition2}
Hernandez-Ortega, J., Galbally, J., Fierrez, J., Beslay, L.: Biometric quality:
  Review and application to face recognition with faceqnet. arXiv preprint
  arXiv:2006.03298  (2020)

\bibitem{gan_development4}
Hudson, D.A., Zitnick, L.: Generative adversarial transformers. In:
  International Conference on Machine Learning. pp. 4487--4499. PMLR (2021)

\bibitem{bidirectional_gan}
Jaiswal, A., AbdAlmageed, W., Wu, Y., Natarajan, P.: Bidirectional conditional
  generative adversarial networks. In: Asian Conference on Computer Vision. pp.
  216--232. Springer (2018)

\bibitem{debias_adversarial_forgetting_yz}
Jaiswal, A., Moyer, D., Ver~Steeg, G., AbdAlmageed, W., Natarajan, P.:
  Invariant representations through adversarial forgetting. In: Proc. of the
  AAAI Conf. on Artificial Intelligence. vol.~34, pp. 4272--4279 (2020)

\bibitem{gan_development1}
Jiang, Y., Chang, S., Wang, Z.: Transgan: Two pure transformers can make one
  strong gan, and that can scale up. Advances in Neural Information Processing
  Systems  \textbf{34} (2021)

\bibitem{preprocessing1}
Kamiran, F., Calders, T.: Data preprocessing techniques for classification
  without discrimination. Knowledge and Information Systems  \textbf{33}(1),
  1--33 (2012)

\bibitem{FairFace}
Karkkainen, K., Joo, J.: Fairface: Face attribute dataset for balanced race,
  gender, and age for bias measurement and mitigation. In: Proc. of the
  IEEE/CVF Winter Conf. on Applications of Computer Vision. pp. 1548--1558
  (2021)

\bibitem{StyleGAN3}
Karras, T., Aittala, M., Laine, S., H{\"a}rk{\"o}nen, E., Hellsten, J.,
  Lehtinen, J., Aila, T.: Alias-free generative adversarial networks. In:
  Beygelzimer, A., Dauphin, Y., Liang, P., Vaughan, J.W. (eds.) Advances in
  Neural Information Processing Systems (2021),
  \url{https://openreview.net/forum?id=Owggnutk6lE}

\bibitem{styleGan2}
Karras, T., Laine, S., Aittala, M., Hellsten, J., Lehtinen, J., Aila, T.:
  Analyzing and improving the image quality of stylegan. In: Proc. of the
  IEEE/CVF Conf. on Computer Vision and Pattern Recognition. pp. 8110--8119
  (2020)

\bibitem{debias4_extreme_bias_MI_colored_MNIST}
Kim, B., Kim, H., Kim, K., Kim, S., Kim, J.: Learning not to learn: Training
  deep neural networks with biased data. In: Proc. of the IEEE/CVF Conf. on
  Computer Vision and Pattern Recognition. pp. 9012--9020 (2019)

\bibitem{DECAF}
Kyono, T., van Breugel, B., Berrevoets, J., van~der Schaar, M.: Decaf:
  Generating fair synthetic data using causally-aware generative networks.
  NeurIPS. cc  (2021)

\bibitem{faderNet}
Lample, G., Zeghidour, N., Usunier, N., Bordes, A., Denoyer, L., Ranzato, M.:
  Fader networks: Manipulating images by sliding attributes. Advances in neural
  information processing systems  \textbf{30} (2017)

\bibitem{debias3_ba2}
Leino, K., Fredrikson, M., Black, E., Sen, S., Datta, A.: Feature-wise bias
  amplification. In: International Conf. on Learning Representations (2019),
  \url{https://openreview.net/forum?id=S1ecm2C9K7}

\bibitem{RLB}
Li, J., Abd-Almageed, W.: Information-theoretic bias assessment of learned
  representations of pretrained face recognition. In: 2021 16th IEEE
  International Conference on Automatic Face and Gesture Recognition (FG 2021).
  pp.~1--8. IEEE (2021)

\bibitem{negative_positive}
Li, X., Jia, X., Jing, X.Y.: Negative-aware training: Be aware of negative
  samples. In: ECAI 2020, pp. 1269--1275. IOS Press (2020)

\bibitem{debias_resampling_representation_bias_colored_MNIST}
Li, Y., Vasconcelos, N.: Repair: Removing representation bias by dataset
  resampling. In: Proc. of the IEEE/CVF Conf. on Computer Vision and Pattern
  Recognition. pp. 9572--9581 (2019)

\bibitem{debias_cost_sensitive}
Ling, C.X., Sheng, V.S.: Cost-sensitive learning and the class imbalance
  problem. Encyclopedia of machine learning  \textbf{2011},  231--235 (2008)

\bibitem{FID_benchmark_CelebA_64}
Liu, L., Ren, Y., Lin, Z., Zhao, Z.: Pseudo numerical methods for diffusion
  models on manifolds. In: International Conference on Learning Representations
  (2022), \url{https://openreview.net/forum?id=PlKWVd2yBkY}

\bibitem{stgan}
Liu, M., Ding, Y., Xia, M., Liu, X., Ding, E., Zuo, W., Wen, S.: Stgan: A
  unified selective transfer network for arbitrary image attribute editing. In:
  Proceedings of the IEEE/CVF conference on computer vision and pattern
  recognition. pp. 3673--3682 (2019)

\bibitem{CelebA}
Liu, Z., Luo, P., Wang, X., Tang, X.: Deep learning face attributes in the
  wild. In: Proc. of the IEEE International Conf. on computer vision. pp.
  3730--3738 (2015)

\bibitem{FID_comparable1}
Maggipinto, M., Terzi, M., Susto, G.A.: Introvac: Introspective variational
  classifiers for learning interpretable latent subspaces. Engineering
  Applications of Artificial Intelligence  \textbf{109},  104658 (2022)

\bibitem{IJBC}
Maze, B., Adams, J., Duncan, J.A., Kalka, N., Miller, T., Otto, C., Jain, A.K.,
  Niggel, W.T., Anderson, J., Cheney, J., et~al.: Iarpa janus benchmark-c: Face
  dataset and protocol. In: 2018 International Conf. on Biometrics (ICB). pp.
  158--165. IEEE (2018)

\bibitem{SensitiveNets}
Morales, A., Fierrez, J., Vera-Rodriguez, R., Tolosana, R.: Sensitivenets:
  Learning agnostic representations with application to face images. IEEE
  Transactions on Pattern Analysis and Machine Intelligence  \textbf{43}(6),
  2158--2164 (2020)

\bibitem{gan_development2}
Niemeyer, M., Geiger, A.: Giraffe: Representing scenes as compositional
  generative neural feature fields. In: Proceedings of the IEEE/CVF Conference
  on Computer Vision and Pattern Recognition. pp. 11453--11464 (2021)

\bibitem{FID_comparable2}
Pandey, K., Mukherjee, A., Rai, P., Kumar, A.: Diffusevae: Efficient,
  controllable and high-fidelity generation from low-dimensional latents. arXiv
  preprint arXiv:2201.00308  (2022)

\bibitem{gan_development5}
Patashnik, O., Wu, Z., Shechtman, E., Cohen-Or, D., Lischinski, D.: Styleclip:
  Text-driven manipulation of stylegan imagery. In: Proceedings of the IEEE/CVF
  International Conference on Computer Vision. pp. 2085--2094 (2021)

\bibitem{race2}
Pezdek, K., Bland{\'o}n-Gitlin, I., Moore, C.: Children's face recognition
  memory: more evidence for the cross-race effect. The J. of applied psychology
   \textbf{88 4},  760--3 (2003)

\bibitem{HSIC}
Quadrianto, N., Sharmanska, V., Thomas, O.: Discovering fair representations in
  the data domain. In: Proceedings of the IEEE/CVF Conference on Computer
  Vision and Pattern Recognition. pp. 8227--8236 (2019)

\bibitem{hat_glasses_correlation}
Ramaswamy, V.V., Kim, S.S., Russakovsky, O.: Fair attribute classification
  through latent space de-biasing. In: Proceedings of the IEEE/CVF Conference
  on Computer Vision and Pattern Recognition. pp. 9301--9310 (2021)

\bibitem{balanced_dataset_BFW}
Robinson, J.P., Livitz, G., Henon, Y., Qin, C., Fu, Y., Timoner, S.: Face
  recognition: too bias, or not too bias? In: Proceedings of the IEEE/CVF
  Conference on Computer Vision and Pattern Recognition Workshops. pp.~0--1
  (2020)

\bibitem{gan_augment_real_data3}
Sandfort, V., Yan, K., Pickhardt, P.J., Summers, R.M.: Data augmentation using
  generative adversarial networks (cyclegan) to improve generalizability in ct
  segmentation tasks. Scientific reports  \textbf{9}(1), ~1--9 (2019)

\bibitem{generate_fair2_fairness_GAN}
Sattigeri, P., Hoffman, S.C., Chenthamarakshan, V., Varshney, K.R.: Fairness
  gan: Generating datasets with fairness properties using a generative
  adversarial network. IBM Journal of Research and Development
  \textbf{63}(4/5), ~3--1 (2019)

\bibitem{generate_fair3}
Sharmanska, V., Hendricks, L.A., Darrell, T., Quadrianto, N.: Contrastive
  examples for addressing the tyranny of the majority. arXiv preprint
  arXiv:2004.06524  (2020)

\bibitem{vgg16}
Simonyan, K., Zisserman, A.: Very deep convolutional networks for large-scale
  image recognition. arXiv preprint arXiv:1409.1556  (2014)

\bibitem{def_minority_bias}
Stone, R.S., Ravikumar, N., Bulpitt, A.J., Hogg, D.C.: Epistemic
  uncertainty-weighted loss for visual bias mitigation. In: Proceedings of the
  IEEE/CVF Conference on Computer Vision and Pattern Recognition. pp.
  2898--2905 (2022)

\bibitem{inceptionv3}
Szegedy, C., Vanhoucke, V., Ioffe, S., Shlens, J., Wojna, Z.: Rethinking the
  inception architecture for computer vision. In: Proceedings of the IEEE
  conference on computer vision and pattern recognition. pp. 2818--2826 (2016)

\bibitem{correlation_distance}
Sz{\'e}kely, G.J., Rizzo, M.L., Bakirov, N.K., et~al.: Measuring and testing
  dependence by correlation of distances. The annals of statistics
  \textbf{35}(6),  2769--2794 (2007)

\bibitem{End}
Tartaglione, E., Barbano, C.A., Grangetto, M.: End: Entangling and
  disentangling deep representations for bias correction. In: Proceedings of
  the IEEE/CVF conference on computer vision and pattern recognition. pp.
  13508--13517 (2021)

\bibitem{FR_bias_study}
Terh{\"o}rst, P., Kolf, J.N., Huber, M., Kirchbuchner, F., Damer, N., Moreno,
  A.M., Fierrez, J., Kuijper, A.: A comprehensive study on face recognition
  biases beyond demographics. IEEE Transactions on Technology and Society
  \textbf{3}(1),  16--30 (2021)

\bibitem{alignment_good2}
Thom, N., Hand, E.M.: Facial attribute recognition: A survey. Computer Vision:
  A Reference Guide pp. 1--13 (2020)

\bibitem{directional_bias_amplification}
Wang, A., Russakovsky, O.: Directional bias amplification. arXiv preprint
  arXiv:2102.12594  (2021)

\bibitem{RL-RBN}
Wang, M., Deng, W.: Mitigate bias in face recognition using skewness-aware
  reinforcement learning. arXiv preprint arXiv:1911.10692  (2019)

\bibitem{debias_domain_discriminative_RFW_MI_AUC_TAR_FAR}
Wang, M., Deng, W., Hu, J., Tao, X., Huang, Y.: Racial faces in the wild:
  Reducing racial bias by information maximization adaptation network. In:
  Proc. of the IEEE/CVF International Conf. on Computer Vision. pp. 692--702
  (2019)

\bibitem{gan_development6}
Wang, X., Xie, L., Dong, C., Shan, Y.: Real-esrgan: Training real-world blind
  super-resolution with pure synthetic data. In: Proceedings of the IEEE/CVF
  International Conference on Computer Vision. pp. 1905--1914 (2021)

\bibitem{debias_domain_independent_training_BA_usage}
Wang, Z., Qinami, K., Karakozis, I.C., Genova, K., Nair, P., Hata, K.,
  Russakovsky, O.: Towards fairness in visual recognition: Effective strategies
  for bias mitigation. In: Proc. of the IEEE/CVF Conf. on Computer Vision and
  Pattern Recognition. pp. 8919--8928 (2020)

\bibitem{FairGAN}
Xu, D., Yuan, S., Zhang, L., Wu, X.: Fairgan: Fairness-aware generative
  adversarial networks. In: 2018 IEEE International Conference on Big Data (Big
  Data). pp. 570--575. IEEE (2018)

\bibitem{fairgan+}
Xu, D., Yuan, S., Zhang, L., Wu, X.: Fairgan+: Achieving fair data generation
  and classification through generative adversarial nets. In: 2019 IEEE
  International Conference on Big Data (Big Data). pp. 1401--1406. IEEE (2019)

\bibitem{debias_transfer_learning}
Yin, X., Yu, X., Sohn, K., Liu, X., Chandraker, M.: Feature transfer learning
  for face recognition with under-represented data. In: Proceedings of the
  IEEE/CVF Conference on Computer Vision and Pattern Recognition. pp.
  5704--5713 (2019)

\bibitem{debias2_ACC}
Zemel, R., Wu, Y., Swersky, K., Pitassi, T., Dwork, C.: Learning fair
  representations. In: Proc. of the 30th International Conf. on International
  Conf. on Machine Learning - Volume 28. p. III–325–III–333. ICML'13,
  JMLR.org (2013)

\bibitem{synthetic_data_face_recognition}
Zhang, H., Grimmer, M., Ramachandra, R., Raja, K., Busch, C.: On the
  applicability of synthetic data for face recognition. In: 2021 IEEE
  International Workshop on Biometrics and Forensics (IWBF). pp.~1--6. IEEE
  (2021)

\bibitem{UTKFace}
Zhang, Z., Song, Y., Qi, H.: Age progression/regression by conditional
  adversarial autoencoder. In: Proc. of the IEEE Conf. on computer vision and
  pattern recognition. pp. 5810--5818 (2017)

\bibitem{debias_age}
Zhao, J., Yan, S., Feng, J.: Towards age-invariant face recognition. IEEE
  Transactions on Pattern Analysis and Machine Intelligence  (2020)

\bibitem{Image_caption_Bias_amplification_yz}
Zhao, J., Wang, T., Yatskar, M., Ordonez, V., Chang, K.W.: Men also like
  shopping: Reducing gender bias amplification using corpus-level constraints.
  In: Proceedings of the 2017 Conference on Empirical Methods in Natural
  Language Processing. pp. 2941--2951 (2017),
  \url{https://www.aclweb.org/anthology/D17-1319}

\bibitem{cross_sample_MI_minimization}
Zhu, W., Zheng, H., Liao, H., Li, W., Luo, J.: Learning bias-invariant
  representation by cross-sample mutual information minimization. In:
  Proceedings of the IEEE/CVF International Conference on Computer Vision. pp.
  15002--15012 (2021)

\bibitem{Long_tail}
{Zhu}, X., {Anguelov}, D., {Ramanan}, D.: Capturing long-tail distributions of
  object subcategories. In: 2014 IEEE Conf. on Computer Vision and Pattern
  Recognition. pp. 915--922 (2014). \doi{10.1109/CVPR.2014.122}

\bibitem{gan_augment_real_data2_long_tail}
Zhu, X., Liu, Y., Li, J., Wan, T., Qin, Z.: Emotion classification with data
  augmentation using generative adversarial networks. In: Pacific-Asia
  conference on knowledge discovery and data mining. pp. 349--360. Springer
  (2018)

\end{thebibliography}
}

\appendix
\addcontentsline{toc}{section}{Appendices}
\renewcommand{\thesubsection}{\Alph{subsection}}
\section*{Appendix}

We include additional studies of our method in the following sections.

\subsection{Difference from face-attribute manipulation.}
Our work first applies face synthesis methods (\eg StyleGAN2) to systematically and methodically synthesize images to construct high-quality fair dataset to theoretically satisfy the fairness criteria rather than solely propose a pipeline to synthesize the facial attributes dataset. FaderNet~\cite{faderNet} and STGAN~\cite{stgan} manipulate face attributes to generate different realistic face images based on GANs. Different from~\cite{faderNet, stgan} which take image generation as the main objective, we focus on fairer dataset construction.

\subsection{Comparison with different backbones.}
We follow the same experiment setup in Section 4.4 to train a ResNet-50 as backbone to classify all non-sex-related AOIs and evaluate the classification and fairness performance trained with other backbones including ResNet-101~\cite{ResNet50}, Inception-v3~\cite{inceptionv3} and VGG-16~\cite{vgg16}. The mAP over all backbones is $73.1\pm0.2$. For the representative of two kinds of bias scores, KL is $0.16\pm0.02$ and RLB is $0.65\pm0.05$. The small variations demonstrate our proposed synthetic dataset is model-agnostic.

\subsection{Discussion on the size of synthetic dataset.}
\begin{figure}[htbp]
\centering
\begin{minipage}[b]{.24\linewidth}
    \centering
    \subfloat[][{Accuracy.}]{\label{fig:acc}\includegraphics[width=0.9\linewidth]{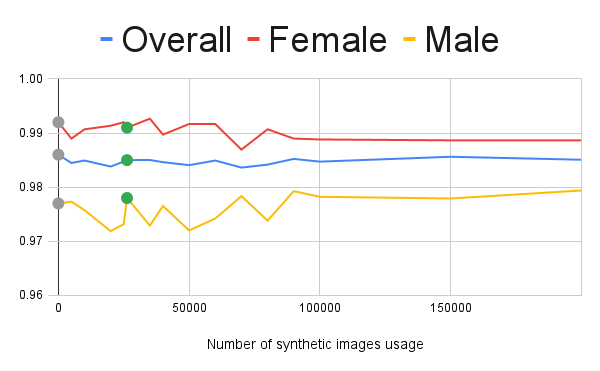}}
\end{minipage}
\begin{minipage}[b]{.24\linewidth}
    \centering
    \subfloat[][{BA.}]{\label{fig:ba}\includegraphics[width=0.9\linewidth]{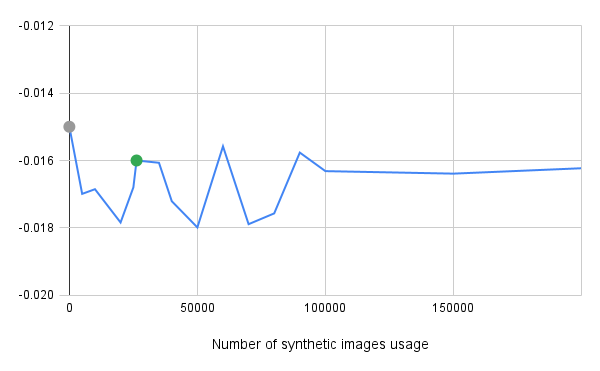}}
\end{minipage}
\begin{minipage}[b]{.24\linewidth}
    \centering
    \subfloat[][{${dcor}^2$.}]{\label{fig:dcor}\includegraphics[width=0.9\linewidth]{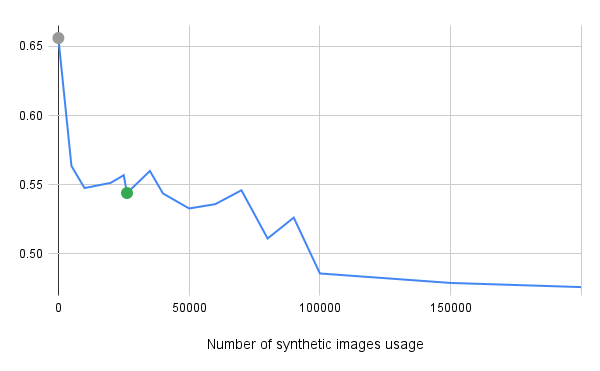}}
\end{minipage}
\begin{minipage}[b]{.24\linewidth}    
  \centering
    \subfloat[][RLB.]{\label{fig:RLB}\includegraphics[width=.9\linewidth]{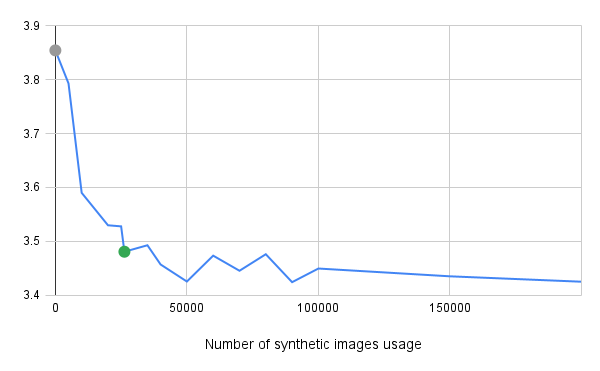}}
\end{minipage}
    \caption{The change of performance as the number of synthetic dataset increases. The grey point represents results under the original imbalanced dataset training, and the green point represents results under \texttt{Supplement} settings where we use $26248$ synthetic male images to supplement the lack in the original dataset.}
\label{fig:upperbound}
\end{figure}

In this section, we further discuss the relation between the number of synthetic images usage and performance on sex classification. As shown in~\cref{fig:upperbound}, fairness has been improved as the number of synthetic training images increases. Although BA converges in the beginning, the other two metrics based on \emph{statistical dependence} decrease quickly before green point, which demonstrates skew of original dataset harms fairness, and finally converge at the lower bound as more training images are used. The lower bound of these metrics infers the upper bound of fairness, which is the best performance of our method. Since there are two main challenges with respect to \emph{dataset bias} and \emph{task bias}, and our method is based on the facial attributes-level balance to address \emph{dataset bias}, the remaining bias may stem from the lack of generalized representation, \ie \emph{task bias}, which should be resolved by debiasing models. Specifically, after training with facial attribute-level balanced dataset, the recognition difficulty in the minority demographic group may be due to the intrinsic difficulty from specialty of facial attributes instead of imbalanced training.

\subsection{Choice of GAN Backbone}
We use StyleGAN2~\cite{styleGan2} as our facial image generator. Even compared with StyleGAN3~\cite{StyleGAN3} where style mixing is not naturally supported, style mixing is one of the main strengths of StyleGAN2, which requires less time to train and yields better style mixing images. On the other hand, StyleGAN3 is an Alias-Free GAN with \emph{translation/rotation equivalence} without the need to align the training dataset, which is less important to our proposed methods compared to style mixing since Aligned \& Cropped subset of CelebA dataset~\cite{CelebA} is available and has been demonstrated to be of good quality~\cite{alignment_good1,alignment_good2}.

\subsection{Attributes Study}
In Section 4.1, we study the nature of \emph{dataset bias} in CelebA dataset at the facial attribute level. In this section, we supplement more details of the experiment and results.

\noindent
\textbf{Experiment Setup.}
We train separate attribute classifiers with ResNet-50~\cite{ResNet50} to recognize 39 facial attributes on the sex-level balanced CelebA dataset sampled from original CelebA dataset. Together with Average Precision (AP), we present the positive sample rate among female images and male images for each attribute. Difference in Equal Opportunity (DEO) is calculated by the absolute value of the difference between AP among females and AP among males. We make statistics based on 50 trials of random choice to balance the origin dataset at the sex level.

\noindent
\textbf{Results.}
As shown in~\cref{tab:attribute_study}, we can find that the remaining bias even after sex-level balanced training is due to the distribution difference of facial attributes instead of apparently sex difference. Based on listed positive sample rate and AP, we can find a roughly proportional relationship between the number of positive samples and AP. More specifically, neural networks tend to learn discriminative representations from positive samples instead of negative samples~\cite{negative_positive}. However, although positive samples are more easily learned for neural networks than negative samples, there are insufficient positive samples of some specific attributes in the minority domain. Thus, we propose a pipeline to generate synthesized datasets with both protected attribute-level balance and facial attribute-level balance to effectively fix this gap by mitigating the facial attributes or appearance difference across sex.

\begin{table}[htbp]
\centering
 \caption{\emph{Dataset bias} exists in sex-level balanced CelebA dataset by long tail distribution in the real world.}
\label{tab:attribute_study}
\resizebox{0.84\textwidth}{!}{%
\begin{tabular}{cccccccc}
\toprule
Attribute   Type       & \multicolumn{3}{c}{Positive Rate (\%)} & \multicolumn{3}{c}{AP}  & Fairness \\ \cmidrule(lr){1-1} \cmidrule(){2-4} \cmidrule(l){5-7} \cmidrule(lr){8-8} 
Masculinity Attributes & Female      & Male      & Overall      & Female & Male & Overall & DEO $\downarrow$                \\ \midrule
5'o clock shadow       & 0.0         & 27.0      & 13.5         & -      & 82.1 & 82.1    & -                   \\
Bald                   & 0.0         & 5.0       & 2.5          & -      & 84.5 & 84.5    & -                   \\
Bushy Eyebrows         & 7.0         & 24.0      & 15.5         & 55.6   & 85.1 & 76.7    & 29.5                \\
Goatee                 & 0.0         & 15.0      & 7.5          & -      & 80.5 & 80.5    & -                   \\
Mustache               & 0.0         & 10.0      & 5.0          & -      & 65.0 & 65.0    & -                   \\
Receding Hairline      & 5.0         & 12.0      & 8.5          & 58.8   & 64.3 & 61.7    & 5.5                 \\
Sideburns              & 0.0         & 13.0      & 6.5          & -      & 86.2 & 86.2    & -                   \\
Wearing necktie        & 0.0         & 17.0      & 8.5          & -      & 73.3 & 73.3    & -                        \\ \midrule
Femininity Attributes   & Female      & Male      & Overall      & Female & Male & Overall & DEO $\downarrow$                  \\ \midrule
Arched Eyebrows        & 42.0        & 5.0       & 23.5         & 77.6   & 47.6 & 75.8    & 30.0                \\
Attractive             & 68.1        & 28.0      & 48.0         & 94.0   & 72.4 & 91.2    & 21.6                \\
Heavy makeup           & 66.0        & 0.0       & 33.0         & 96.7   & -    & 96.7    & -                   \\
No beard               & 100.0       & 61.0      & 80.5         & 100.0  & 97.9 & 99.8    & 2.0                 \\
Oral Face              & 33.0        & 22.0      & 27.5         & 65.4   & 43.4 & 59.5    & 22.0                \\
Rosy cheeks            & 11.0        & 0.0       & 5.5          & 68.7   & -    & 68.7    & -                   \\
Wearing Earrings       & 31.0        & 2.0       & 16.5         & 86.0   & 46.8 & 84.8    & 39.2                \\
Wearing lipstick       & 80.1        & 1.0       & 40.6         & 99.0   & 22.1 & 98.9    & 76.9                \\
Wearing Necklace       & 19.9        & 2.0       & 11.0         & 41.3   & 13.6 & 40.4    & 27.8                  \\ \midrule
Unbiased Attributes   & Female      & Male      & Overall      & Female & Male & Overall & DEO $\downarrow$                  \\ \midrule
Bangs                  & 20.0        & 8.0       & 14.0         & 94.4   & 89.6 & 93.4    & 4.8                 \\
Big lips               & 30.0        & 16.0      & 23.0         & 55.6   & 60.6 & 58.2    & 4.9                 \\
Blurry                 & 5.0         & 6.0       & 5.5          & 66.8   & 61.8 & 64.6    & 5.0                 \\
Eyeglasses             & 2.0         & 12.0      & 7.0          & 98.9   & 98.3 & 98.4    & 0.6                 \\
Mouth slightly open    & 52.3        & 42.0      & 47.2         & 98.8   & 98.1 & 98.6    & 0.7                 \\
Narrow Eyes            & 11.0        & 12.0      & 11.5         & 54.7   & 53.6 & 54.2    & 1.0                 \\
Smiling                & 54.0        & 40.0      & 47.0         & 98.8   & 96.8 & 98.3    & 2.0                 \\
Wearing Hat            & 2.9         & 8.0       & 5.4          & 91.0   & 95.8 & 94.4    & 4.8                 \\
Young                  & 88.0        & 64.0      & 76.0         & 97.8   & 93.3 & 96.8    & 4.5             \\ \midrule
Attribute of Interest  & Female      & Male      & Overall      & Female & Male & Overall & DEO $\downarrow$                  \\ \midrule
Black Hair             & 20.0        & 29.0      & 24.5         & 86.3   & 92.8 & 89.0    & 6.5                 \\
Blond Hair             & 24.0        & 2.0       & 13.0         & 92.6   & 58.7 & 91.2    & 33.9                \\
Brown Hair             & 24.0        & 15.0      & 19.5         & 77.3   & 70.2 & 75.4    & 7.1                 \\
Chubby                 & 1.0         & 12.0      & 6.5          & 36.3   & 62.0 & 58.5    & 25.7                \\
Gray Hair              & 1.0         & 9.0       & 5.0          & 62.7   & 79.4 & 76.8    & 16.7                \\
Pale skin              & 6.0         & 2.0       & 4.0          & 68.2   & 46.6 & 64.1    & 21.6                \\
Straight Hair          & 19.0        & 24.0      & 21.5         & 61.7   & 67.7 & 64.0    & 6.0                 \\
Wavy Hair              & 45.0        & 14.0      & 29.5         & 90.0   & 66.8 & 87.1    & 23.2                \\
High cheekbones        & 56.0        & 31.0      & 43.5         & 97.2   & 84.4 & 94.8    & 12.8                \\
Bags under Eyes        & 10.0        & 35.0      & 22.5         & 52.7   & 67.0 & 62.8    & 14.3                \\
Big nose               & 10.0        & 42.0      & 26.0         & 46.4   & 75.9 & 68.4    & 29.5                \\
Double chin            & 1.0         & 10.0      & 5.5          & 27.3   & 61.7 & 57.2    & 34.4                \\
Pointy nose            & 36.0        & 16.0      & 26.0         & 66.2   & 43.8 & 61.8    & 22.3                 \\ \bottomrule
\end{tabular}%
}
\end{table}

\subsection{Taxonomy of Facial Attributes in CelebA Dataset}
In Section 4.1, we summarize all 39 facial attributes in CelebA dataset into three groups. In this section, we present more details for each group.

First, \emph{unbiased attributes} are the facial attributes which do not yield much bias (\ie DEO is less than 5 as shown in~\cref{tab:attribute_study}), \eg \textsf{Bangs}, \textsf{Big Lips}, \textsf{Blurry}, \textsf{Eyeglasses}, \textsf{Mouth Slightly Open}, \textsf{Narrow Eyes}, \textsf{Smiling}, \textsf{Wearing Hat}, \textsf{Young}. To reserve resources for the biased facial attributes, \emph{unbiased attributes} may be secondary to be balanced at the facial attribute level.

Besides, \emph{masculinity/femininity attributes} are considered as \emph{Attribute of Interest} (AOI) in Section 4.3 (sex classification). Specifically, \emph{masculinity attributes} include \textsf{5'o Clock Shadow}, \textsf{Bald}, \textsf{Bushy Eyebrows}, \textsf{Goatee}, \textsf{Mustache}, \textsf{Receding Hairline}, \textsf{Sideburns}, \textsf{Wearing Necktie}, and \emph{femininity attributes} include \textsf{Arched Eyebrows}, \textsf{Heavy Makeup}, \textsf{No Beard}, \textsf{Oral Face}, \textsf{Rosy Cheeks}, \textsf{Wearing Earrings}, \textsf{Wearing Lipstick}, \textsf{Wearing Necklace}, \textsf{Attractive}. 

Finally, we categorize the attributes which are not sex-related but induce \emph{dataset bias} even with sex-level balanced training as AOI appending \emph{masculinity/femininity attributes} in Section 4.4 (facial attribute classification), \eg \textsf{Black Hair}, \textsf{Blond Hair}, \textsf{Brown Hair}, \textsf{Chubby}, \textsf{Gray Hair}, \textsf{Pale Skin}, \textsf{Straight Hair}, \textsf{Wavy Hair}, \textsf{High Cheekbones}, \textsf{Bags Under Eyes}, \textsf{Big Nose}, \textsf{Double Chin}, \textsf{Pointy Nose}.

\subsection{Mutually Exclusive Attributes}
In Section 4.4, we present one representative facial attribute for the mutually exclusive facial attributes. In this section, we provide the additional results of other facial attributes. 

As shown in~\cref{tab:multi_attributes_recognition_mutually_exclusive}, although strategically resampling~\cite{debias_resampling1} outperforms our method under BA, the performances under ${dcor}^2$ and RLB (which are more consistent and stable as pointed by~\cite{RLB}) are not good. Furthermore, with better DEO and KL, we verify the achievement of \emph{equal opportunity} and \emph{equalized odds} in Section 3, which is not achievable for sex-level balanced dataset and strategically resampling.

\begin{table}[htbp]
\centering
\caption{Performance and fairness comparison on facial attribute recognition.}
\label{tab:multi_attributes_recognition_mutually_exclusive}
\resizebox{1\textwidth}{!}{%
\begin{tabular}{ccccccc}
\toprule
                      &               & GrayHair       & BlackHair      & BrownHair      & StraightHair   & Average        \\
\cmidrule(lr){3-3} \cmidrule(lr){4-4} \cmidrule(lr){5-5} \cmidrule(lr){6-6} \cmidrule(lr){7-7} 
\multirow{4}{*}{AP $\uparrow$}   & Baseline      & \textbf{78.8}  & \textbf{90.6}  & \textbf{77.1}  & \textbf{66.7}  & \textbf{78.3}  \\  
                      & Resampling~\cite{debias_resampling1}   & 71.8           & 72.3           & 72.0           & 66.2           & 70.6           \\
                      & GAN-Debiasing~\cite{hat_glasses_correlation} & 77.9           & 88.1           & 74.1           & 57.7           & 74.4           \\
                      & Ours          & 78.5           & 89.9           & 75.0           & 64.5           & 77.0           \\ \midrule
\multirow{4}{*}{DEO $\downarrow$} & Baseline      & 25.5           & 6.2            & 8.0            & 6.0            & 11.4           \\
                      & Resampling~\cite{debias_resampling1}    & \textbf{9.6}   & 5.0            & 7.5            & 5.3            & 6.9            \\
                      & GAN-Debiasing~\cite{hat_glasses_correlation} & 25.3           & \textbf{0.8}   & 3.8            & \textbf{5.2}   & 8.8            \\
                      & Ours          & 13.1           & 0.9            & \textbf{2.6}   & 5.5            & \textbf{5.5}   \\ \midrule
\multirow{4}{*}{BA $\downarrow$}   & Baseline      & 0.47           & -0.16          & 0.75           & 1.13           & 0.55           \\
                      & Resampling~\cite{debias_resampling1}    & \textbf{-3.59} & \textbf{-3.51} & \textbf{-3.65} & \textbf{-0.01} & \textbf{-2.69} \\
                      & GAN-Debiasing~\cite{hat_glasses_correlation} & -0.61          & -0.94          & -0.40          & 0.39           & -0.39          \\
                      & Ours          & -0.55          & -0.76          & -1.62          & 0.90           & -0.51          \\ \midrule
\multirow{4}{*}{KL $\downarrow$}   & Baseline      & 0.24           & 0.13           & 0.07           & 0.03           & 0.12           \\
                      & Resampling~\cite{debias_resampling1}    & \textbf{0.11}  & 0.10           & 0.07           & \textbf{0.02}  & 0.08           \\
                      & GAN-Debiasing~\cite{hat_glasses_correlation} & 0.23           & 0.03           & 0.04           & 0.03           & 0.08           \\
                      & Ours          & 0.21           & \textbf{0.02}  & \textbf{0.03}  & 0.03           & \textbf{0.07}  \\ \midrule
\multirow{4}{*}{${dcor}^2 \downarrow$} & Baseline      & 0.67           & 0.30           & 0.31           & 0.39           & 0.42           \\
                      & Resampling~\cite{debias_resampling1}   & 0.38           & 0.28           & 0.30           & 0.30           & 0.31           \\
                      & GAN-Debiasing~\cite{hat_glasses_correlation} & 0.36           & 0.27           & \textbf{0.27}  & \textbf{0.22}  & \textbf{0.28}  \\
                      & Ours          & \textbf{0.34}  & \textbf{0.25}  & \textbf{0.27}  & 0.28           & \textbf{0.28}  \\ \midrule
\multirow{4}{*}{RLB $\downarrow$}  & Baseline      & 0.75           & 0.54           & 0.94           & 1.65           & 0.97           \\
                      & Resampling~\cite{debias_resampling1}    & 0.53           & 0.43           & 0.90           & 0.85           & 0.68           \\
                      & GAN-Debiasing~\cite{hat_glasses_correlation} & \textbf{0.39}  & 0.48           & 0.92           & 1.61           & 0.85           \\
                      & Ours          & 0.56           & \textbf{0.38}  & \textbf{0.67}  & \textbf{0.84}  & \textbf{0.61}  \\ \bottomrule
\end{tabular}
}
\end{table}

\end{document}